\documentclass[journal=jctc,manuscript=article]{achemso}
\SectionNumbersOn
\usepackage[version=3]{mhchem} 
\usepackage[capitalize,noabbrev]{cleveref}
\usepackage{amsmath,amsfonts,bm}
\usepackage{booktabs, multirow}
\usepackage{etoolbox}
\AtBeginDocument{\singlespacing}

\newcommand{\E}{\mathbb{E}}
\newcommand{\R}{\mathbb{R}}
\def\gN{{\mathcal{N}}}
\def\mI{{\bm{I}}}
\def\rmF{{\mathbf{F}}}

\newcommand{\rmx}{\mathbf{x}}
\newcommand{\rmw}{\mathbf{w}}
\newcommand{\rmh}{\mathbf{h}}

\newcommand{\fg}{\rmx} 
\newcommand{\cg}{\mathbf{z}} 
\newcommand{\cgmap}{\Xi} 
\newcommand{\fpot}{U} 
\newcommand{\cpot}{V} 
\newcommand{\Tau}{\mathcal{T}}
\newcommand{\temp}{k_B \Tau} 

\newcommand{\norm}[1]{\left\lVert #1 \right\rVert}

\usepackage{xspace}

\newcommand*{\ie}{{\it i.e.}\@\xspace}

\makeatletter
\patchcmd{\acs@contact@details}{E}{*\,E}{}{}
\makeatother

\author{Marloes Arts}
\affiliation[internship]{Work done during an internship at Microsoft Research (Amsterdam).}
\alsoaffiliation[KU]{University of Copenhagen, Department of Computer Science, Universitetsparken 1, Copenhagen, 2100, Denmark.}
\altaffiliation{Equal contribution.}
\email{ma@di.ku.dk}

\author{Victor Garcia Satorras}
\email{victorgar@microsoft.com}
\altaffiliation{Equal contribution.}
\author{Chin-Wei Huang}
\affiliation[MSR-A]{AI4Science, Microsoft Research, Evert van de Beekstraat 354, Amsterdam, 1118 CZ, The Netherlands.}

\author{Daniel Zügner}
\affiliation[MSR-B]{AI4Science, Microsoft Research, Karl-Liebknecht-Stra{\ss}e 32, Berlin, 10178, Germany.}

\author{Marco Federici}
\affiliation[internship]{Work done during an internship at Microsoft Research (Amsterdam).}
\alsoaffiliation[amlab]{University of Amsterdam, Informatics Institute, Science Park 904, Amsterdam, 1098 XH, The Netherlands.}

\author{Cecilia Clementi}
\affiliation[MSR-B]{AI4Science, Microsoft Research, Karl-Liebknecht-Stra{\ss}e 32, Berlin, 10178, Germany.}
\alsoaffiliation[FUBerlin]{Freie Universit\"at Berlin, Department of Physics, Arnimalle 12, Berlin, 14195, Germany.}

\author{Frank No\'e}
\affiliation[MSR-B]{AI4Science, Microsoft Research, Karl-Liebknecht-Stra{\ss}e 32, Berlin, 10178, Germany.}

\author{Robert Pinsler}
\affiliation[MSR-C]{AI4Science, Microsoft Research, 21 Station Road, Cambridge, CB1 2FB, United Kingdom.}

\author{Rianne van den Berg}
\affiliation[MSR-A]{AI4Science, Microsoft Research, Evert van de Beekstraat 354, Amsterdam, 1118 CZ, The Netherlands.}

\title{Two for One: Diffusion Models and Force Fields for Coarse-Grained Molecular Dynamics}

\begin{document}







\begin{abstract}
Coarse-grained (CG) molecular dynamics enables the study of biological processes at temporal and spatial scales that would be intractable at an atomistic resolution. However, accurately learning a CG force field remains a challenge. In this work, we leverage connections between score-based generative models, force fields and molecular dynamics to learn a CG force field without requiring any force inputs during training. Specifically, we train a diffusion generative model on protein structures from molecular dynamics simulations, and we show that its score function approximates a force field that can directly be used to simulate CG molecular dynamics. While having a vastly simplified training setup compared to previous work, we demonstrate that our approach leads to improved performance across several protein simulations for systems up to 56 amino acids, reproducing the CG equilibrium distribution, and preserving dynamics of all-atom simulations such as protein folding events.
\end{abstract}


\newpage
\section{Introduction}

\begin{figure}[!b]
\centering
    \includegraphics[width=0.4\textwidth]{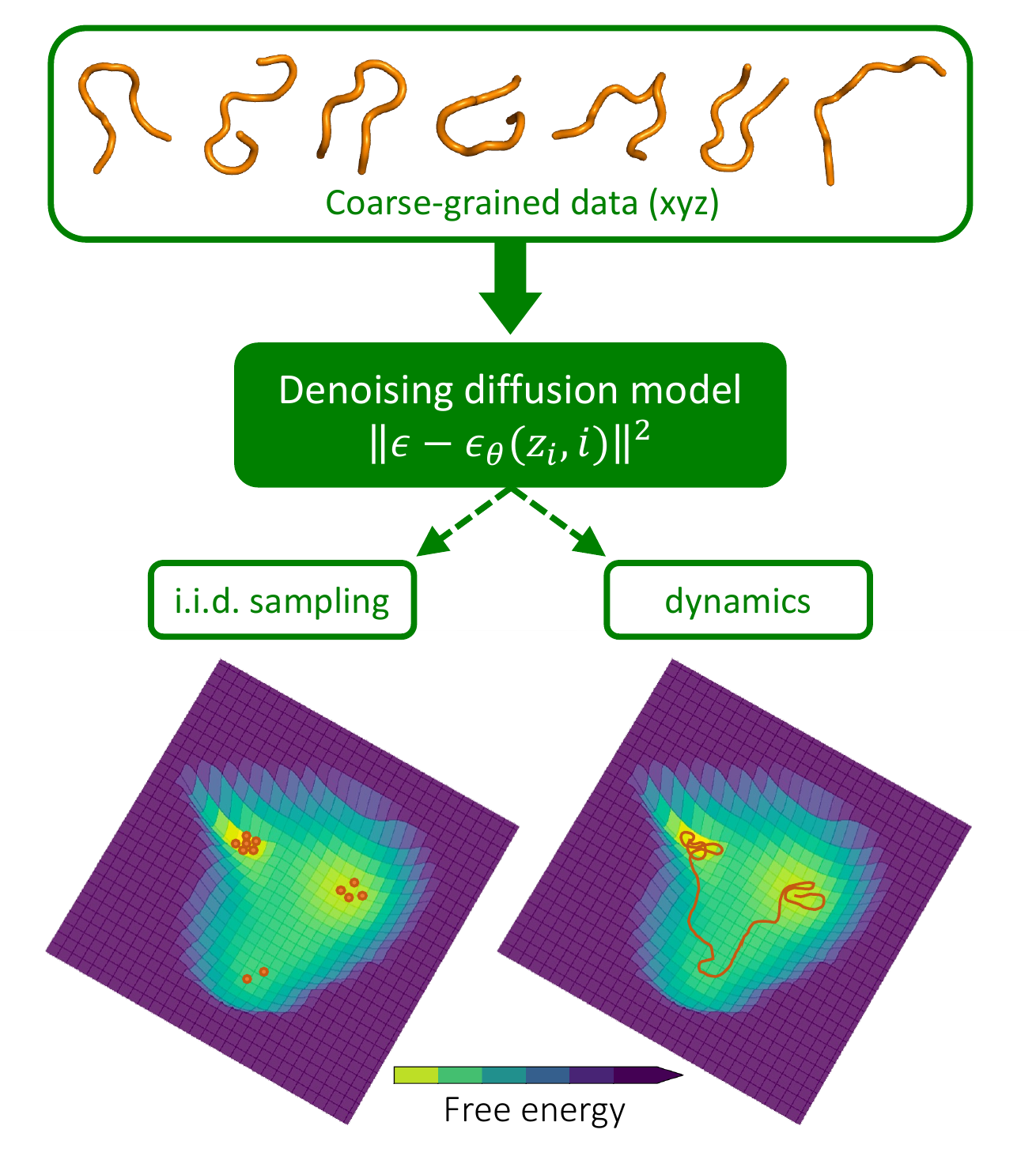}
    \caption{A denoising diffusion model is trained with a standard loss on atomistic (fine-grained) equilibrium samples projected onto the CG space. By leveraging connections between score-based generative modeling, force fields and molecular dynamics, we obtain a single model that can generate i.i.d. equilibrium CG samples \textit{and} whose neural network can be used as a CG force field in CG molecular dynamics simulations.}
    \label{fig:introfigure}
\end{figure}

Coarse-grained (CG) molecular dynamics (MD) promises to scale simulations to larger spatial and time scales than currently accessible through atomistic MD simulations \citep{clementiCOSB2008,noid2013perspective, saunders2013coarse, kmiecik2016coarse}. Scaling up MD by orders of magnitude would enable new studies on macromolecular dynamics over longer ranges of time, such as large protein folding events and slow interactions between large molecules.

To obtain a CG simulation model, one first maps the all-atom, or \textit{fine-grained}, representation to a \textit{coarse-grained} representation, e.g. by grouping certain atoms together to form so-called CG beads.
Second, a CG force field needs to be designed such that CG molecular dynamics simulations reproduce relevant features of molecular systems.

In top-down approaches, a CG model is often defined to reproduce specific macroscopic observables, as experimentally measured and/or simulated on fine-grained models 
\citep{marrink2007martini,davtyan2012awsem,Matysiak2006,Chen2018}. 
In bottom-up approaches, one seeks to obtain a CG model reproducing the microscopic behavior (e.g., thermodynamics, kinetics) of a fine-grained model \citep{noid2008multiscale, shell2008relative,Nueske2019}. In the latter case, a common approach is to define a CG force field for the chosen CG representation by enforcing thermodynamic consistency \citep{noid2013perspective}. This requires that simulations following the CG model should have the same equilibrium distribution as obtained by projecting equilibrated all-atom simulations onto the CG resolution.

Traditional bottom-up coarse-graining techniques that rely on the thermodynamic consistency principle have produced significant results in the last decade \cite{Mim2012,Chu2005,Yu2021}, in particular when used in combination with machine-learning methods \citep{wang2019machine,husic2020coarse}. 
Two commonly used approaches are variational force matching and relative entropy minimization. 

Variational force matching minimizes the mean squared error between the model's CG forces and the atomistic forces projected onto the CG space, which must be included in the data \citep{noid2008multiscale}. However, due to the stochastic nature of the projected forces, this noisy force-matching estimator has a large variance, leading to data-inefficient training.  
Alternatively, relative entropy minimization approaches \citep{shell2008relative} perform density estimation in the CG space without accessing atomistic forces. The majority of this class of methods are equivalent to energy-based models \citep{song2021train}. Since training these models requires iteratively drawing samples from the model to estimate log-likelihood gradients, such methods demand significantly higher computational cost \citep{hinton2002training}.

Flow-matching \citep{kohler2022force} is a hybrid approach that does not require atomistic forces for training (like relative entropy minimization) while also retaining good sample-efficiency.
The method has two training stages. First, a CG density is modeled with an augmented normalizing flow \citep{rezende2015variational, papamakarios2021normalizing, huang2020augmented, chen2020vflow}. 
A second learning stage with a force-matching-like objective is then required to extract a deterministic CG force field that can be used in CG molecular dynamics simulations. \citet{kohler2022force} demonstrated that flow-matching improves performance on several fast-folding proteins \citep{lindorff2011fast}. However, the learned CG models are not yet accurate enough for reproducing the thermodynamics of the corresponding fine-grained models, and scaling to larger proteins leads to instabilities.

In this work, we leverage the recently popularized class of denoising diffusion models \citep{ho2020denoising, sohl2015deep}, which have already shown promising results for protein and molecular structure generation \citep{wu2022protein, trippe2022diffusion, watson2022broadly, igashov2022equivariant, qiao2022dynamic, hoogeboom2022equivariant}, conformer generation \citep{jing2022torsional}, and docking \citep{corso2022diffdock}. In particular, we train a score-based generative model on CG structures sampled from the CG equilibrium distribution. By highlighting connections between score-based generative models \citep{song2020score}, force fields and molecular dynamics, we demonstrate that learning such a generative model with a standard denoising loss and a conservative score yields a single model that can be used to produce i.i.d. CG samples \textit{and} which can be used directly as a CG force field for CG molecular dynamics simulations. An overview is shown in \cref{fig:introfigure}.
In addition to having a single-stage training setup, our method leads to improved performance across several protein simulations for systems up to 56 amino acids, reproducing the CG equilibrium distribution, and preserving the dynamical mechanisms observed in all-atom simulations such as protein folding events. We also provide evidence that our diffusion CG model allows for scaling to a larger protein than previously accessible through flow-matching.

\section{Background}

Coarse-graining can be described by a dimensionality reduction map $\cgmap:\R^{3N}\rightarrow\R^{3n}$ that transforms a high-dimensional atomistic representation $\fg \in \R^{3N}$ in 3D space to a lower-dimensional CG representation $\cg\in\R^{3n}$, where $n \ll N$. 
For molecular systems, the CG map is usually linear, $\cgmap \in \R^{3n\times 3N}$, and returns the Cartesian coordinates $\cg$ of CG ``beads'' as a linear combination of the Cartesian coordinates $\fg$ of a set of representative atoms.

The probability density of the atomistic system at a particular temperature $\mathcal{T}$ is described by the Boltzmann distribution $q(\fg) \propto \exp\left(-U(\fg)/\temp\right)$, where $U(\fg)$ is the system's potential energy and $k_B$ is the Boltzmann constant.
By identifying the ensemble of atomistic configurations $\fg$ that maps into the same CG configuration $\cg$, we can explicitly express the probability density of the CG configurations $\cg$ as:
\begin{align} 
    q(\cg) = \frac{\int \exp(-\fpot(\fg)/\temp) \delta(\cgmap(\fg)-\cg) \,d\fg}{\int \exp(-\fpot(\fg')/\temp)d\,\fg'},
    \label{eq:boltzmann_marginalized}
\end{align}
where $\delta(\cdot)$ is the Dirac delta function. 
Up to an additive constant, this distribution uniquely defines the thermodynamically consistent effective CG potential of mean force $\cpot(\cg)$  \citep{noid2008multiscale}:
\begin{align*}
    \cpot(\cg) &= -\temp \log{q(\cg)} + \mathrm{cst.} \\
    &= -\temp \log{\int e^{- \fpot(\fg)/\temp} \delta(\cgmap(\fg)-\cg) \,d\fg} + \mathrm{cst.}
\end{align*}
Unfortunately, computing the integral is usually intractable. Therefore, methods that approximate thermodynamically consistent effective CG potentials have been proposed. Below we briefly summarize two commonly used approaches.

\paragraph{Variational force matching.}
\citet{noid2008multiscale} showed that under certain constraints of the coarse-graining mapping $\cgmap$, a more tractable consistency equation between the coarse-grained force field $-\nabla_\cg \cpot(\cg)$ and the atomistic force field $-\nabla_\fg \fpot(\fg)$ can be obtained. More specifically, if $\cgmap$ is a linear map, and if each bead has at least one atom with a nonzero coefficient only for that specific bead, then the following relation holds: $-\nabla_\cg \cpot(\cg) = \E_{q(\fg \mid \cg)}[\cgmap_f(-\nabla_\fg \fpot(\fg))]$.
Here, $\cgmap_f$ is a linear map whose coefficients are related to the linear coefficients of the CG map $\cgmap$ \citep{Ciccotti2005}. \citet{noid2008multiscale} showed that the above relation can be used to approximate a thermodynamically consistent CG potential $\cpot_\theta(\cg)$ with parameters $\theta$ by minimizing the following variational loss:
\begin{align}
    \E_{q(\fg,\cg)}\left[\norm{ \nabla_\cg \cpot_\theta(\cg) - \cgmap_f(\nabla_\fg \fpot(\fg)) }_2^2 \right].
    \label{eq:varforce_loss}
\end{align}

\paragraph{Relative entropy minimization.}
Another approach to obtaining the CG forces is via relative entropy minimization, where optimizing the density implicitly leads to optimized mean potential functions. 
Concretely, we seek to estimate the CG density by minimizing the relative entropy, or Kullback-Leibler divergence, 
$\E_{q(\cg)}[\log q(\cg)-\log p_\theta(\cg)]$, which is equivalent to optimizing the maximum likelihood when a finite number of samples is drawn from $q(\cg)$.
The approximate CG forces can be extracted from the optimized model density $p_\theta(\cg)$ through $-\nabla_\cg \cpot_\theta(\cg) \propto \nabla_\cg \log p_\theta(\cg)$.
Unlike variational force matching, relative entropy minimization does not impose any constraints on the CG map, and no atomistic forces are required for training.

Traditionally, an unnormalized version of $p_\theta$ is modeled by directly parameterizing the CG potential $\cpot_\theta$, yielding $p_\theta(\cg)\propto \exp\left(-\cpot_\theta(\cg)/\temp\right)$.
To minimize the relative entropy, one would need to either estimate the free energy (\ie the normalizing constant) of the model \citep{shell2008relative} or draw i.i.d. samples from the model for gradient estimation \citep{hinton2002training, thaler2022deep}, which renders this approach impractical for higher-dimensional problems.

An alternative is to use an explicit density model such as a normalizing flow \citep{dinh2014nice, rezende2015variational, dinh2016density}
, allowing for straightforward maximum-likelihood density estimation and force field learning. However, learning expressive invertible functions is challenging, so
instead, \citet{kohler2022force} opted for augmented normalizing flows \citep{huang2020augmented, chen2020vflow}. The introduction of auxiliary random variables increases the expressivity of the flow, at the cost of an intractable marginal likelihood, yielding a minimization objective that is a variational upper bound to the relative entropy. Furthermore, one can only extract a stochastic estimate for the CG force from the augmented normalizing flow model. In order to distill a deterministic approximate CG force to simulate the CG dynamics, \citet{kohler2022force} proposed a teacher-student setup akin to variational force-matching. This two-stage approach was dubbed flow-matching.

\section{Diffusion Models for Coarse-Grained Molecular Dynamics}
\label{sec:diffusion-model}

Denoising diffusion probabilistic models (DDPMs) \citep{ho2020denoising, sohl2015deep} sample from a probability distribution by approximating the inverse of a diffusion process, \ie a denoising process. The diffusion (forward) process is defined as a Markov chain of $L$ steps $q(\cg_{1:L}\mid\cg_0) = \prod_{i=1}^L q(\cg_i \mid \cg_{i-1})$, where $\cg_0$ is a sample from the unknown data distribution $q(\cg_0)$. The learned reverse process is defined as a reverse-time Markov chain of $L$ denoising steps $p(\cg_{0:L}) := p(\cg_L)\prod_{i=1}^L p_\theta(\cg_{i-1} \mid \cg_{i})$ that starts from the prior $p(\cg_L)$.
For real-valued random variables, the choice of distribution for the forward process is typically Gaussian, $q(\cg_i \mid \cg_{i-1}) = \gN(\cg_i ; \sqrt{1-\beta_i} \cg_{i-1}, \beta_i \mI)$, with $\{\beta_i\}$ predetermined variance parameters that increase as a function of $i$ such that the Markov chain has a standard Normal stationary distribution. The reverse process distributions are chosen to have the same functional form: $p(\cg_L)=\gN(\mathbf{0}, \mI)$ and $p_\theta(\cg_{i-1} \mid \cg_{i}) = \gN(\cg_{i-1} ; \mu_\theta(\cg_{i}, i), \sigma_i^2 \mI)$. Here, $\mu_\theta(\cg_i,i)$ is a learnable function with parameters $\theta$, and $\sigma_i^2$ is a fixed variance for noise level $i$ that is determined by $\beta_i$. By making use of closed-form marginalization for Gaussian distributions, and by parameterizing the means as $\mu_\theta(\cg_i, i) = \frac{1}{\sqrt{\alpha_i}}\left(\cg_i - \frac{\beta_i}{\sqrt{1-\bar{\alpha}_i}}\epsilon_\theta(\cg_i, i)\right)$, with $\epsilon_\theta(\cg_i, i)$ the noise prediction neural network, training proceeds by minimizing the loss \citep{ho2020denoising}:
\begin{equation}
    \label{eq:elbo_noise}  
     \sum_{i=1}^L  K_i \E_{q(\cg_0)}\E_{\mathcal{N(\epsilon;\mathbf 0 , \mathbf I)}} \left[ \norm{\epsilon \!- \!\epsilon_\theta(\sqrt{\bar{\alpha}_i}\cg_0 + \sqrt{1  -  \bar{\alpha}_i}\epsilon, i)}^2 \right].
\end{equation}
Here, $\alpha_i=1-\beta_i$ and $\bar{\alpha}_i=\prod_{s=1}^i\alpha_s$.
Up to a constant, Eq.~\ref{eq:elbo_noise} is a negative evidence lower bound if $K_i=\frac{\beta_i^2}{2\sigma^2_i \alpha_i(1-\bar{\alpha}_i)}$. However, \citet{ho2020denoising} found that a reweighted loss with $K_i=1$ worked best in practice. 

In this manuscript, the data consists of samples from the CG Boltzmann distribution: $q(\cg_0) \propto e^{-\frac{V(\cg)}{\temp}}$. Given a trained diffusion model parameterized through a noise prediction network $\epsilon_\theta(\cg_i,i)$, we can produce i.i.d. samples of the approximate CG distribution through ancestral sampling from the graphical model $p(\cg_L)\prod_{i=1}^L p_{\theta}(\cg_{i-1} \mid \cg_i)$.

\subsection{Extracting Force Fields from Diffusion Models} \label{sec:method_force_field}

\citet{song2020score} demonstrated that the DDPM loss in Eq.~\ref{eq:elbo_noise} with $K_i=1$ is equivalent to the following weighted sum of denoising score matching objectives \citep{vincent2011connection}:
\begin{equation} 
\sum_{i=1}^L (1\! - \! \bar{\alpha}_i) \E_{q(\cg_0)} \E_{q(\cg_i\mid \cg_0)}\! \left[\norm{s_{\theta}(\cg_i, i) \! - \! \nabla_{\cg_i} \! \log q(\cg_i\mid \cg_0) }^2\right].
\label{eq:denoising_score_matching}
\end{equation}
Here, $q(\cg_i\mid \cg_0) = \gN(\cg_i ; \sqrt{\bar{\alpha}_i} \cg_0, (1-\bar\alpha_i) \mI)$, and $s_{\theta}(\cg_i, i)$ is the score model. 
While this was not made explicit by \citet{song2020score}, the equivalence of these two losses is achieved by relating the score model $s_\theta(\cg_i,i)$ to the noise prediction network $\epsilon_\theta(\cg_i,i)$ through $s_{\theta}(\cg_i, i) = -\frac{\epsilon_\theta(\cg_i,i)}{\sqrt{1-\bar\alpha_i}}$, see supplementary information section A.1.
Given a sufficiently expressive model and sufficient amounts of data, the optimal score $s_{\theta^*}(\cg_i, i)$ will match the score $\nabla_{\cg_i}\log q(\cg_i)$ \citep{vincent2011connection}, 
where $q(\cg_i) = \int \mathrm d \cg_0 q(\cg_i\mid \cg_0) q(\cg_0)$ is the marginal distribution at level $i$ of the forward diffusion process. At sufficiently low noise levels, the marginal distribution $q(\cg_i)$ will resemble the data distribution $q(\cg_0)$, such that $s_{\theta^*}(\cg_i,i)$ effectively approximates the score of the unknown data distribution. When the latter is equal to the CG Boltzmann distribution $q(\cg_0)\propto e^{-\frac{V(\cg)}{\temp}}$,
the optimal score $s_{\theta^*}(\cg_i, i)$ at level $i=1$ will approximately match the CG forces $\nabla_\cg \log q(\cg) = \frac{-\nabla_\cg V(\cg)}{\temp} = \frac{\rmF_\cg}{\temp}$. 
Finally, by using the relation between $s_\theta(\cg_i,i)$ and the noise prediction network $\epsilon_\theta(\cg_i,i)$, we can extract the approximate CG forces from a denoising diffusion model trained with the loss in Eq.~\ref{eq:elbo_noise}:
\begin{align}  \label{eq:cg_force_field}
    \mathbf{F}_\cg^{DFF} = -\frac{\temp}{\sqrt{1-\bar{\alpha}_i}}\mathbf{\epsilon}_{\theta^*}(\cg, i).
\end{align}
We will refer to such an approximate CG force field as a Denoising Force Field (DFF).
While in principle the lowest level ($i=1$) should provide the best approximation to the CG forces, in practice, we treat $i$ as a hyperparameter and pick the best $i$ by cross-validating the simulated dynamics. 

Connections between force fields and denoising diffusion models have been made in previous work. \citet{zaidi2022pre} pre-trained a property prediction graph neural network in a denoising diffusion setup by denoising molecular structures that locally maximize the Boltzmann distribution (or minimize the energy).
By approximating the data distribution as a mixture of Gaussians centered around these local minima, they demonstrate that the score matching objective is equivalent to learning the force field of this approximate mixture of Gaussians data distribution.
Similarly, \citet{xie2021crystal} connected the learned score in a denoising network for small noise levels to a harmonic force field around energy local minima structures. A key point is that these connections only provide approximate force fields around the local minima structures, making them of limited use in downstream tasks. In this work, we show that training denoising diffusion models on samples from the equilibrium Boltzmann distribution---rather than only the locally maximizing structures---allows us to learn an approximate force field in an unsupervised manner for the entire equilibrium distribution. This is crucial for running stable and reliable CG MD simulations with the extracted CG force field.

\subsection{Molecular Dynamics with the Denoising Force Field} \label{sec:molecular_dynamics}
With the DFF from Eq.~\ref{eq:cg_force_field}, we can perform CG molecular dynamics simulations by propagating the Langevin equation
\begin{equation} \label{eq:cg_langevin_dynamics}
    M\frac{\mathrm d^2 \cg}{\mathrm d\; t^2} = -\nabla_\cg V(\cg)   - \gamma M\frac{\mathrm d\cg}{\mathrm d t} + \sqrt{2 M \gamma \temp }\rmw(t),
\end{equation}
where we substitute $-\nabla_\cg V(\cg) =  \mathbf{F}_\cg^{DFF}$. $M$ represents the mass of the CG beads, $\gamma$ is a friction coefficient, and $\rmw(t)$ is a delta-correlated stationary Gaussian process $\E_{p(\fg)} \left[\rmw(t)  \cdot \rmw(t') \right] = \delta(t-t')$ with mean $\E_{p(\fg)} \left[ \rmw(t)\right]   = 0$.    
In our experiments, we set $\gamma$ and $\mathcal T$ to the same values as those used in the original atomistic simulations that produced the data. Therefore, given a trained network $\epsilon_\theta$, the only hyper-parameter left to tune is the noise level $i$. Further information regarding the trade-off of noise level $i$ can be found in supplementary information section B.3.

A well-known limit of the Langevin equation (Eq.~~\ref{eq:cg_langevin_dynamics}) is that of a negligible mass and a large friction coefficient (with a finite $\eta = \gamma M $), called Brownian dynamics or overdamped Langevin dynamics. Interestingly, in supplementary information section A.2 we show that iteratively diffusing and denoising at a low-noise level (e.g. $i=1$) approximates Brownian dynamics with a simulation timestep $\Delta t$ implicitly defined through $ \Delta t \frac{\temp}{M \gamma} = 1-\bar\alpha_1 = \beta_1$.

\subsection{Denoising Force Field Architecture} \label{sec:denoising_network}

The choice of the neural network $\epsilon_\theta$ is heavily influenced by the physical symmetries of the system under study. For instance, the CG force field must be conservative, i.e. it must equal the negative gradient of a CG energy potential $V_\theta(\cg)$. 
Therefore, we parameterize $\epsilon_\theta(
\cg_i, i)$ as the gradient of an energy neural network with a scalar output, \ie $\epsilon_\theta(
\cg_i, i) = \nabla_{\cg_i} \mathrm{nn}_\theta(\cg_i, i)$, with $\mathrm {nn}_\theta: \mathbb R^{3n}\times \{1, ..., L\} \mapsto \mathbb R$.  Previous studies on image generation by \citet{salimans2021should} yielded no empirical difference in sample quality when using an unconstrained score network or a score that is parameterized as the gradient of an energy function. However, in supplementary information section B.1 we demonstrate that using a conservative score in a diffusion model is crucial for stable CG MD simulations with the extracted denoising force field.

Furthermore, the force field must be translation invariant and rotation equivariant.
We ensure the model is translation invariant by using the coordinates of the CG beads only through pairwise difference vectors $\cg_{(i)} - \cg_{(j)}$ as input to the network.
While the forces must be equivariant to rotations 
,we explicitly do \textit{not} want reflection equivariance, to avoid generating mirrored proteins as reported by \citet{trippe2022diffusion}. In other words, our goal is to achieve equivariance with respect to SO(3) instead of O(3), as opposed to other works that use relative distances as the input representation \citep{satorras2021n}. 
A simple strategy to approximate SO(3) equivariance without requiring the more expensive spherical harmonics or angular representations is the use of data augmentation. Previous work by \citet{gruver2022lie} showed that learned equivariance with transformers can be competitive with actual equivariant networks. In supplementary information section B.4 we show that our denoising force field learns to be rotation equivariant on a validation set with a relative squared error introduced by rotations of  $< 10^{-6}$.

In this work, we model the network $\mathrm{nn}_\theta$ as a graph transformer adapted with the above symmetry constraints. Further architecture details are given in supplementary information section C.1. Note that previous works on neural-network-based CG force fields also often add a prior energy term in the scalar energy neural network to enforce better behavior of the CG force field further away from the training dataset \citep{wang2019machine, husic2020coarse,kohler2022force}. In contrast, we did not find this to be necessary to obtain stable CG MD simulations with our denoising CG force field.

\section{Experiments}
\label{sec:experiments}

By training our diffusion model on samples from a CG equilibrium distribution, we simultaneously obtain an i.i.d. sample generator (denoted \textit{DFF i.i.d.}) as well as a CG force field for running CG MD simulations (\textit{DFF sim.}). In this section, we evaluate the performance and scalability of our model for both use cases on (i) alanine dipeptide, and (ii) several fast-folding proteins \citep{lindorff2011fast}. In particular, we investigate how well the CG equilibrium distribution and the dynamics can be reproduced.

We compare our model to three baselines: \textit{Flow i.i.d.} and \textit{Flow-CGNet sim.} from \citet{kohler2022force} and \textit{CGNet sim.} \citep{wang2019machine}. \textit{CGNet sim.} is a pure force-matching neural network trained on CG forces that were projected from the fine-grained representation onto the CG representation. \textit{Flow i.i.d.} is the force-agnostic augmented normalizing flow model trained as a density estimator in the first stage of the flow-matching setup \citep{kohler2022force}. This flow model can only be used to produce i.i.d. samples. \textit{Flow-CGNet sim.} performs CG simulations using the deterministic CGNet force field distilled from the gradient of the augmented normalizing flow model in the second teacher-student distillation stage of flow-matching. 
Recall that for our method, we do not require a teacher-student setup since the same network can be used for i.i.d. sampling and for CG simulations. We also provide \textit{reference} data, which is the original MD simulation projected onto the CG resolution.
Lastly, note that while we often show results for both i.i.d. and simulation-based methods, the latter have the more challenging task to also model the dynamics in order to obtain correct equilibrium distributions.
We therefore expect the proposed i.i.d. methods to perform better when analyzing equilibrium distributions.

\subsection{Coarse-Grained Simulation | Alanine Dipeptide}
First, we evaluate our method on a CG representation of the well-studied alanine dipeptide system.
We use the same CG representation as in \citet{kohler2022force, wang2019machine, husic2020coarse}, which projects all atoms onto the five central backbone atoms of the molecule (see supplementary figure S4). The simulated data \citep{kohler2022force} consists of four independent runs of length 500 ns, with 250\,000 samples saved per simulation (2 ps intervals). We evaluate the model using four-fold cross-validation, where three of the simulations are used for training and validation, and one is used for testing.
We consider different training dataset sizes, ranging from $10$K to $500$K training samples.
For the Langevin dynamics simulation, we follow the same settings as \citet{kohler2022force}, \ie we run the simulation at 300 Kelvin for 1M steps with a step size of 2 fs and store the samples every 250 time steps. However, unlike \citet{kohler2022force}, we do not use parallel tempering, which is known to improve the mixing of the dynamics. Our denoising network $\mathrm{nn}_\theta$ (Section \ref{sec:denoising_network}) consists of two graph transformer layers with 96 features in the hidden layers. Further implementation details are in supplementary information section C.4.1.

\paragraph{Metrics.} Following \citet{kohler2022force, wang2019machine}, we evaluate the quality of the generated samples by analyzing statistics over the two dihedral angles $(\phi, \psi)$ computed along the CG backbone of alanine dipeptide. Each angle describes a four-body interaction, representing the main degrees of freedom of the system. We generate a Ramachandran plot by computing the free energy as a function of these two angles, binning values into a 2D histogram, and taking the negative logarithm of the probability density. 
To provide a quantitative analysis, we measure the empirical Jensen-Shannon (JS) divergence between the dihedral distributions of samples drawn from the model and the test set.
A reference comparing the training and test sets is provided as a lower bound.

\begin{figure*}[t] 
    \centering
    \includegraphics[width=1.0\textwidth]{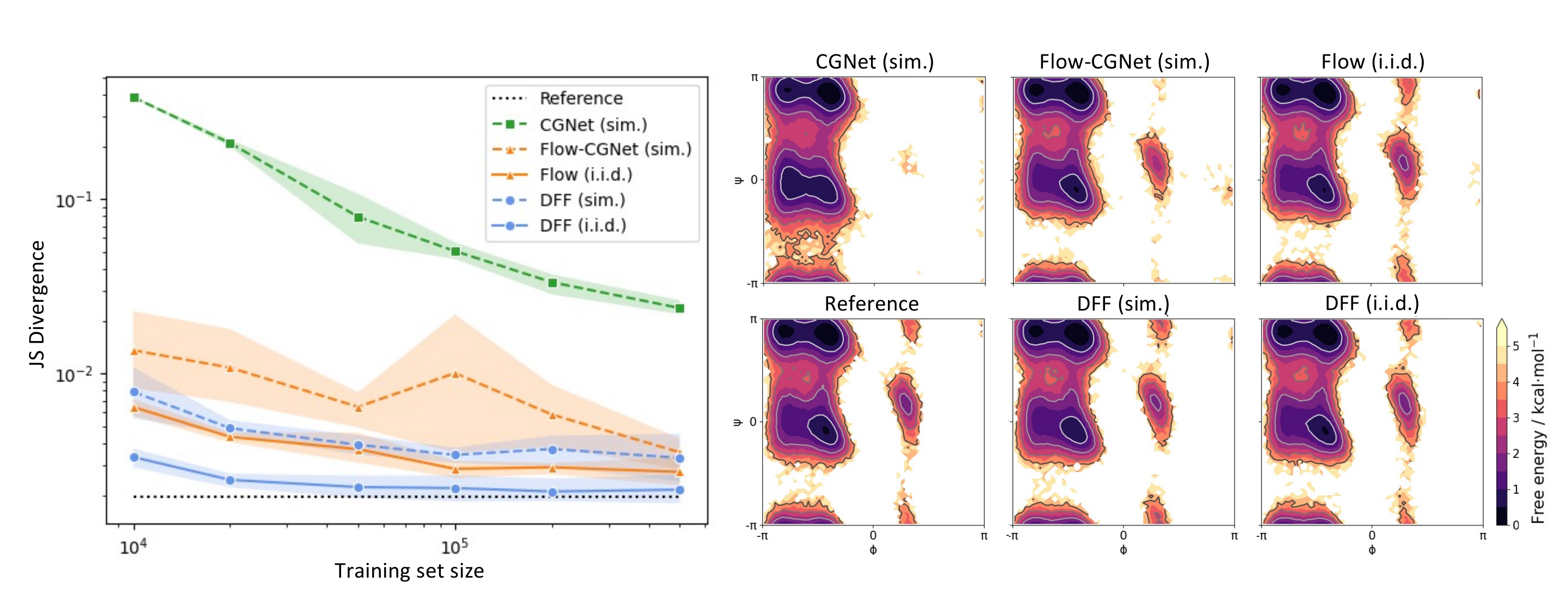}
    \caption{Experimental results for alanine dipeptide. Left: Jensen-Shannon divergence between dihedral distributions produced by several CG methods for different training set sizes and the test partition of the all atom simulation data projected onto the CG resolution. Results are averaged over four runs, and error bars denote a $95\%$ confidence interval. 
    Right: Ramachandran plots showing the dihedral distributions for the different methods trained on 500K samples.}
    \label{fig:alanine}
\end{figure*}

\paragraph{Results.} As shown in \cref{fig:alanine} (left), \textit{DFF sim.} significantly outperforms previous CG simulation methods (\textit{Flow-CGNet sim.} and \textit{CGNet sim.}), especially in the low-data regime, and even performs comparably to the i.i.d. sampling method \textit{Flow i.i.d.}.
Moreover, \textit{DFF i.i.d.} outperforms its counterpart \textit{Flow i.i.d.} with a significant margin, almost approaching the performance of the lower bound (\textit{Reference}).
The right side of \cref{fig:alanine} shows the Ramachandran plots after training on 500 K samples, further highlighting that our model is able to generate realistic samples.

\begin{table*}[t]
\setlength{\tabcolsep}{4pt}
\centering
\caption{Experimental results for fast-folders. The table displays the Jensen-Shannon (JS) divergence for TIC distributions and pairwise distance (PWD) distributions, where in the latter case an average is taken over all entries of the upper triangle of the PWD matrix with offset three. The JS divergences compare distributions from the atomistic MD simulations that were projected on CG space, and the distributions produced by the learned CG methods.}
\label{tab:fastfolder_eq}
\vspace{4pt}
\resizebox{\textwidth}{!}{%
\begin{tabular}{@{}lcrrrrrrrrrr@{}}
\toprule
 &  & \multicolumn{2}{c}{Chignolin} & \multicolumn{2}{c}{Trp-cage} & \multicolumn{2}{c}{Bba} & \multicolumn{2}{c}{Villin} & \multicolumn{2}{c}{Protein G} \\
 &  & \multicolumn{1}{c}{TIC JS} & \multicolumn{1}{c}{PWD JS} & \multicolumn{1}{c}{TIC JS} & \multicolumn{1}{c}{PWD JS} & \multicolumn{1}{c}{TIC JS} & \multicolumn{1}{c}{PWD JS} & \multicolumn{1}{c}{TIC JS} & \multicolumn{1}{c}{PWD JS} & \multicolumn{1}{c}{TIC JS} & \multicolumn{1}{c}{PWD JS} \\ \midrule
Reference &  & .0057 & .0002 & .0026 & .0002 & .0040 & .0002 & .0032 & .0004 & .0014 & .0002 \\ \midrule
Flow & \multirow{2}{*}{i.i.d.} & .0106 & .0022 & .0078 & .0057 & .0229 & .0073 & .0109 & .0142 & n/a & n/a \\
DFF &  & \textbf{.0096} & \textbf{.0005} & \textbf{.0052} & \textbf{.0007} & \textbf{.0111} & \textbf{.0017} & \textbf{.0073} & \textbf{.0009} & .0131 & .0009 \\ \midrule
Flow-CGNet & \multirow{2}{*}{sim.} & .1875 & .1271 & .1009 & .0474 & .1469 & .0594 & .2153 & .0535 & n/a & n/a \\
DFF &  & \textbf{.0335} & \textbf{.0067} & \textbf{.0518} & \textbf{.0403} & \textbf{.1289} & \textbf{.0408} & \textbf{.0564} & \textbf{.0244} & .2260 & .0691 \\ \bottomrule
\end{tabular}%
}
\vspace{-8pt}
\end{table*}

\subsection{Coarse-Grained Simulation | Fast-folding Proteins}
Next, we evaluate our model on a more challenging set of fast-folding proteins \citep{lindorff2011fast}. Such proteins exhibit folding and unfolding events, which makes their simulated trajectories particularly interesting. We pick the same proteins as in \citet{kohler2022force}, namely Chignolin, Trp-cage, Bba and Villin. These were coarse-grained by slicing out the $C_{\alpha}$ atom for every amino acid, yielding one bead per residue (10, 20, 28, and 35 beads, respectively). For these proteins, we produce the \textit{Flow i.i.d.} and \textit{Flow-CGNet sim.} plots through samples that were made publicly available by the authors. Since scaling to larger proteins was found to be challenging for flow-matching \citep{kohler2022force}, we additionally include the larger ``Protein G'' (56 beads) to analyze the scalability of our method. The all atom simulations vary in length, but for each trajectory the frames are shuffled and split 70-10-20\% into a training, validation and test set. More dataset details are given in supplementary information section C.5.1.

\subsubsection{Equilibrium Analysis}

\paragraph{Metrics.} We use several metrics to evaluate the quality of the generated equilibrium distributions.
First, we analyze the slowest changes in the protein conformation, which are usually related to (un-)folding events. For this, we calculate the time-lagged independent component analysis (TICA) \citep{naritomi2011slow, perez2013identification,schwantes2013improvements} using the Deeptime library \cite{hoffmann2021deeptime} and pick the first two TIC coordinates, resulting in a 2D distribution over the slowest processes. Basins in these 2D distributions are associated with meta-stable states. Further, we compute the JS divergence of the obtained TIC distributions between each model and the \textit{reference} MD data (denoted by TIC JS). As a qualitative analysis, we plot the log of the obtained TIC distributions. 

To assess the global structure of the proteins, we compare pairwise distance distributions by calculating the JS divergence relative to the test MD distribution for all distances within the upper triangle of the pairwise distance matrix with a diagonal offset larger than three (denoted by PWD JS). The offset is chosen to avoid over-representing local structure. 
Moreover, we plot the free energy as a function of the root mean squared distance (RMSD) between the generated samples and the native, folded structure for all $\rm{C_{\alpha}}$ atoms. Dips in the resulting curve correspond to meta-stable ensembles with a lower free energy. 
Finally, we analyze the normalized count over contact maps, which results in a 2D histogram that shows the probability of two atoms being in contact, \ie within a threshold of 10 \AA\ of one another.

\begin{figure*}[t]
    \centering
    \includegraphics[width=1\textwidth]{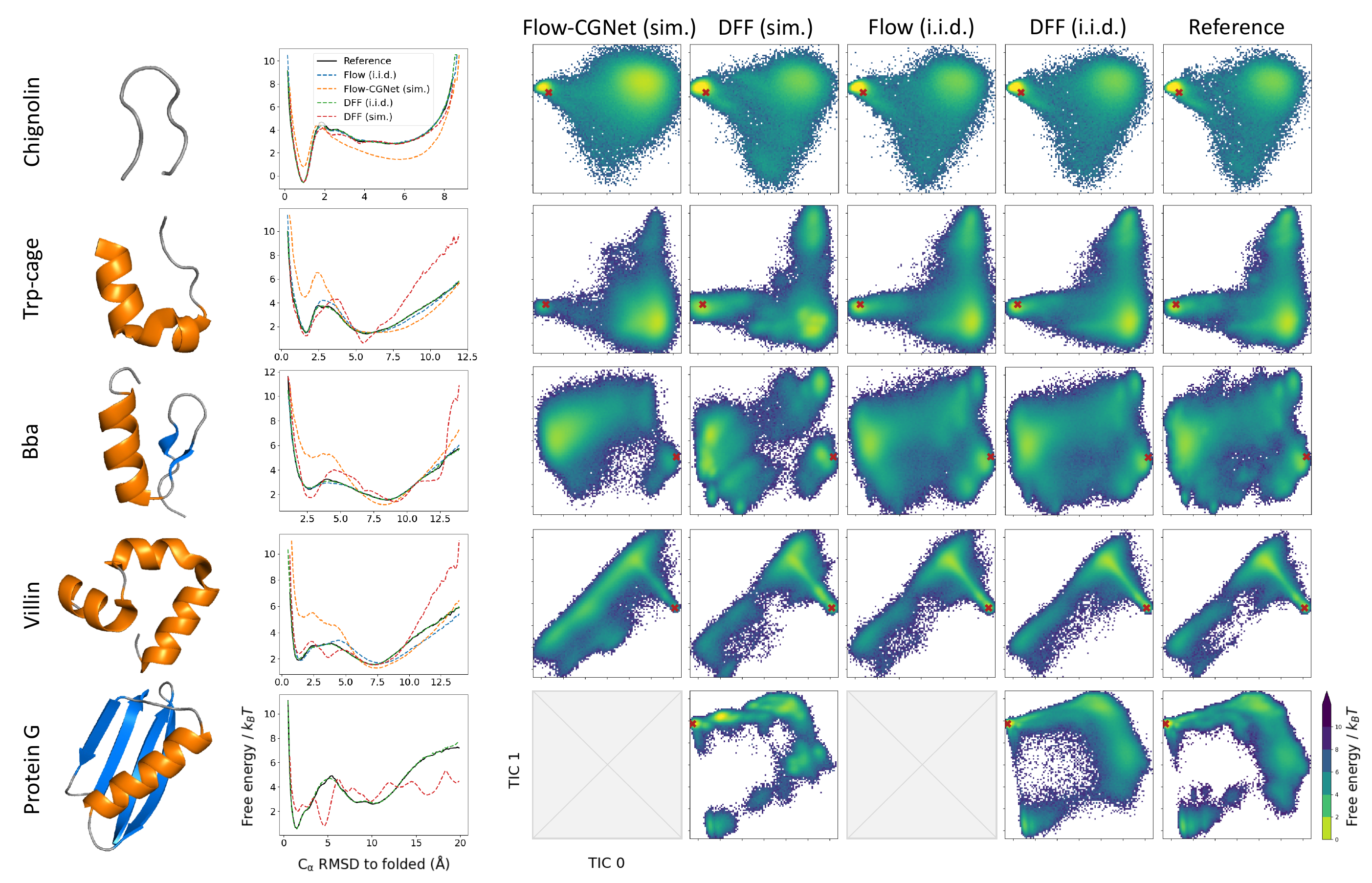}
    \vspace{-10pt}
    \caption{
    Left: native structure visualization with $\rm{\alpha}$-helices in orange and $\rm{\beta}$-sheets in blue. Middle: $\rm{C_{\alpha}}$-RMSD free energy w.r.t. the folded native structure. Right: joint density plots for the two slowest TIC coordinates, where the color indicates the free energy value. The red cross indicates the location of the native structure.}
    \vspace{-10pt}
    \label{fig:tica}
\end{figure*}

\begin{figure}[ht!]
    \begin{center}
    \centerline{\includegraphics[width=0.7\textwidth]{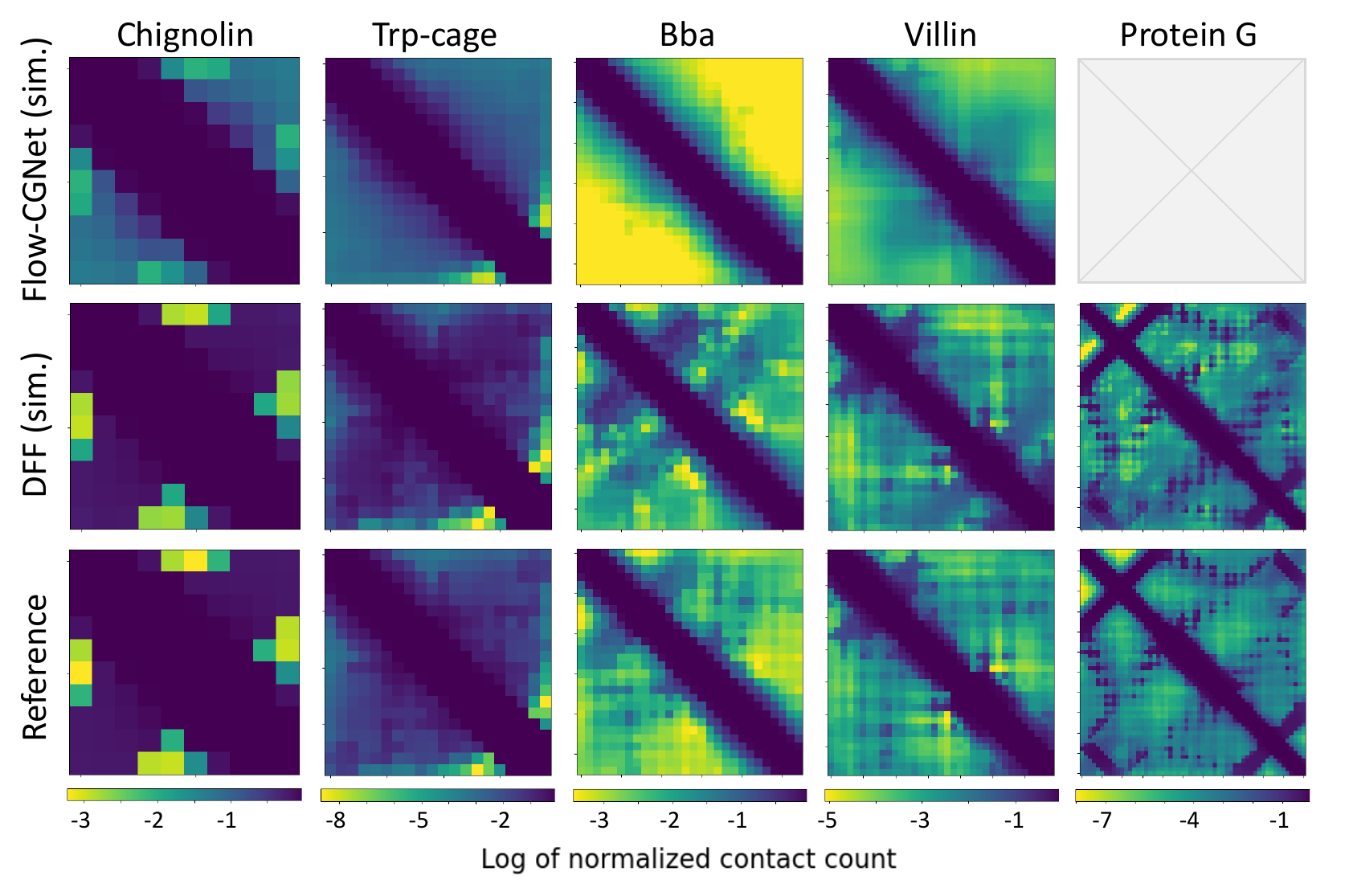}}
    \caption{Contact probability maps for \textit{Flow-CGNet sim.}, \textit{DFF sim.} and the reference MD data for fast-folding proteins. The contact threshold is set to 10 \AA. The axes in the plot represent atom indices. The color indicates the normalized contact count (\ie ``contact probability'') for the corresponding pairwise distance.} 
    \vspace{-10pt}
    \label{fig:contact}
    \end{center}
\end{figure}

\paragraph{Results.} \cref{tab:fastfolder_eq} shows that \textit{DFF i.i.d.} and \textit{DFF sim.} consistently outperform their respective baselines across the equilibrium metrics (TIC JS and PWD JS). As shown in \cref{fig:tica}, the TIC 2D free energy landscapes for our model look more similar to the reference MD distribution (as reflected in the TIC JS metrics), especially for Chignolin and Villin. The similarity is weaker for Bba, where local modes are more dominant. We hypothesize that is because $\beta$-sheets rely stronger on non-local contacts compared to $\alpha$-helices, making them notoriously harder to model. This is particularly challenging in the simulation setting, which is sensitive to the bias-variance tradeoff (as discussed in supplementary information section B.2). Furthermore, the free energy curves as a function of the RMSD are always overlapping with the reference curve for \textit{DFF i.i.d.}, and are close to the reference MD curve in regions with a low free energy for \textit{DFF sim.}. 

\cref{fig:contact} shows further qualitative results in the form of a normalized count over contact maps, \ie ``contact probabilities'', 
for \textit{DFF sim.} and \textit{Flow-CGNet sim.}; for i.i.d. models, see supplementary information section C.5.3. As can be seen, the \textit{DFF} models capture contact probabilities much better than the flow-based models in all proteins, especially in the off-diagonal regions that represent global structure. These results are closely related to the JS divergences between pairwise off-diagonal distances (\cref{tab:fastfolder_eq}). Taken together, these results indicate that diffusion-based models capture global structure better than their flow-based baselines. 
Moreover, the analysis in C.5.3 shows that the CG fast folder samples produced by \textit{DFF sim.} do not display chemical integrity violations such as bond dissociations or backbone crossings, therefore, \textit{DFF sim.} does not require an energy prior as used in \textit{Flow-CGNet sim.} to run stable simulations. Finally, Protein G is a larger and more complex protein compared to the other fast-folders and out of reach for flow-matching models. Our results show that diffusion-based models are scalable to this larger protein and can capture the global structure.

As a limitation in our method, we found that while the \textit{DFF~i.i.d.} model generally improves as we increase the number of features/layers in our neural network, the performance of the simulations obtained by \textit{DFF sim.} is sensitive to the bias/variance trade-off in the network, and it can actually decrease for more flexible networks. See supplementary information section B.2 for an example in Chignolin.

\begin{table*}[t]
\setlength{\tabcolsep}{4pt}
\centering
\caption{Average and state-probability weighted JS divergence between the reference MD data and model simulations for transition probabilities of the estimated Markov model.}
\label{tab:fastfolder_dyn}
\vspace{4pt}
\resizebox{\textwidth}{!}{%
\begin{tabular}{@{}lcccccccc@{}}
\toprule
 & \multicolumn{2}{c}{Chignolin} & \multicolumn{2}{c}{Trp-cage} & \multicolumn{2}{c}{Bba} & \multicolumn{2}{c}{Villin} \\
 & Average & Weighted & Average & Weighted & Average & Weighted & Average & Weighted \\ \midrule
Flow-CGNet & $2.5\cdot10^{-2}$ & $5.7\cdot10^{-3}$ & $4.8\cdot10^{-2}$ & $1.8\cdot10^{-2}$ & $6.9\cdot10^{-2}$ & $6.8\cdot10^{-2}$ & $3.1\cdot10^{-2}$ & $2.9\cdot10^{-2}$ \\
DFF & $\mathbf{9.7\cdot10^{-4}}$ & $\mathbf{5.1\cdot10^{-4}}$ & $\mathbf{1.3\cdot10^{-3}}$ & $\mathbf{7.5\cdot10^{-4}}$ & $\mathbf{4.0\cdot10^{-3}}$ & $\mathbf{4.2\cdot10^{-3}}$ & $\mathbf{1.2\cdot10^{-4}}$ & $\mathbf{2.1\cdot10^{-5}}$ \\ \bottomrule
\end{tabular}%
}
\end{table*}

\begin{figure*}[ht]
    \centering
    \includegraphics[width=1.0\textwidth]{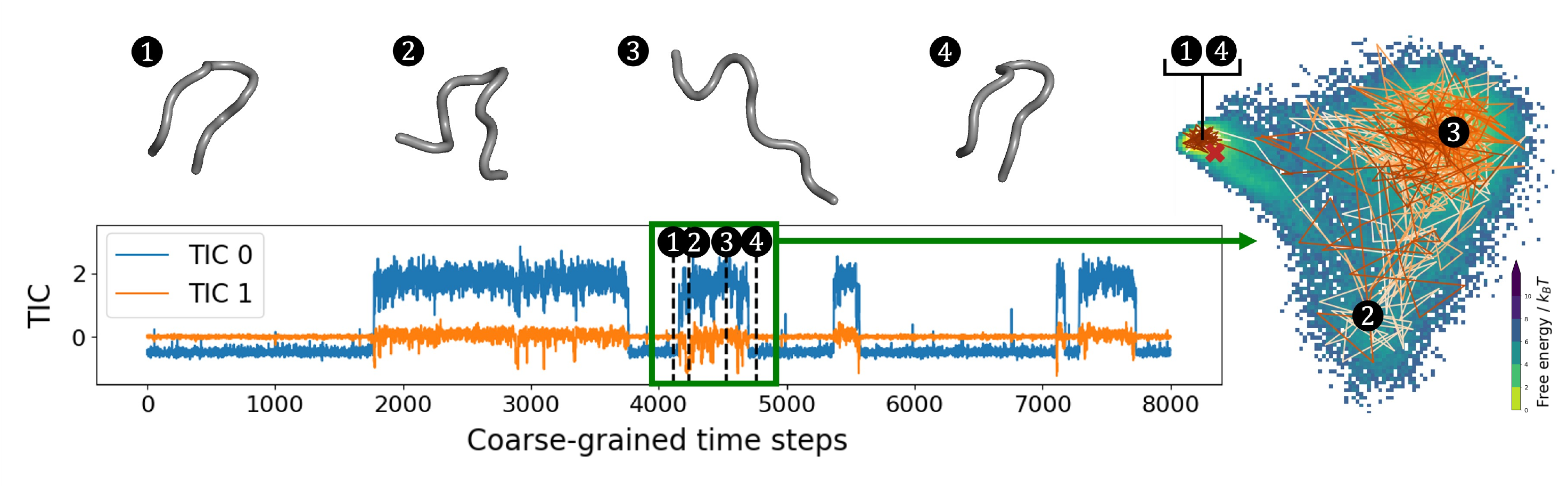}
    \caption{First two TIC coordinates tracked over ``coarse-grained time'' (the CG time step is 2 fs, but there is no direct mapping from coarse- to fine-grained time) for Chignolin, with zoom-in on an unfolding and folding event, showing a 2D trajectory though TIC space and four structures along the path.}
    \label{fig:tictraj}
\end{figure*}

\subsubsection{Dynamics Analysis}
\paragraph{Metrics.} We qualitatively assess the simulated trajectories by tracking the first two TIC coordinates over ``coarse-grained time'' and showing (part of) the corresponding trajectory in 2D TIC space. We visualize (un-)folding events with the corresponding structures along the path. As a quantitative measure, we extract the transition probabilities from one conformational state to the other as follows. First, we use K-means clustering to divide the 2D TIC space into $K$ clusters for the full (unsplit) MD dataset, with $K$ determined by the elbow method. Next, all transitions are counted and normalized to obtain a transition probability matrix corresponding to the estimated Markov model \cite{Prinz2011}, where each row can be compared to MD data using the JS divergence. Even though the relation between fine-grained and coarse-grained time is non-trivial \cite{jin2022understanding,Nueske2019}, leading to different time lags, we can still evaluate how well a coarse-grained model reproduces the kinetic model of the fine-grained reference distribution. We show the average JS divergence over all starting states as well as the average weighted by the overall state probability as estimated from the reference data. Note that this metric can only be calculated for simulation samples (\ie \textit{Flow-CGNet} and \textit{DFF} simulations). Since we compare transition probabilities against the full, unshuffled dataset, there is no test set in this experiment and therefore we cannot calculate a reference value here.

\paragraph{Results.} \cref{fig:tictraj} depicts the first two TIC coordinates for a DFF sim. trajectory in ``coarse-grained time'', clearly showing transitions from the folded to unfolded conformations. This is further highlighted by zooming in on part of the trajectory, revealing how the trajectory moves in 2D TIC space, and what the conformations look like in different parts of the landscape. The results of the transition probability analysis are shown in \cref{tab:fastfolder_dyn} for all fast-folders except Protein G, since no \textit{Flow-CGNet sim.} samples were available for this larger protein. The \textit{DFF sim.} model outperforms the \textit{Flow-CGNet sim.} model across all fast-folders, showing a better preservation of dynamics. More results on transition probability matrices and the clustering of 2D TIC space are in supplementary information section C.5.3.

\section{Conclusions}

We have presented a new approach to CG molecular dynamics modeling based on denoising diffusion models, motivated by connections between score-based generative models, force fields and molecular dynamics. This results in a simple training setup as well as improved performance and scalability compared to previous work.
Future directions to improve our work include 
scaling to larger proteins and generalizing across different systems. Another interesting direction would be to combine the current force-agnostic training approach with an explicit force-matching objective if such force information is available.

\begin{acknowledgement}

The authors express their gratitude to Yaoyi Chen for providing samples and evaluation scripts for the prior work by \citet{kohler2022force}. We also thank Max Welling, Leon Klein, Tor Erlend Fjelde and Andreas Krämer for the insightful discussions and suggestions. 3D molecular visualizations were made using PyMOL \citep{PyMOL}.

\noindent M.A. acknowledges funding from the Novo Nordisk foundation: Center for Basic Machine Learning Research in Life Science (MLLS, grant nr NNF20OC0062606), and project grant (nr NNF18OC0052719). C.C. and F.N. acknowledge funding from the Deutsche Forschungsgemeinschaft DFG (SFB/TRR 186, Project A12; SFB 1114, Projects  A04, B03, and B08; SFB 1078, Project C7; and RTG 2433, Project Q05, NO825/3-2), the National Science Foundation (CHE-1900374, and PHY-2019745), the European Commission (ERC CoG 772230), the Berlin mathematics center MATH+ (AA1-6, AA1-10) and the Einstein Foundation Berlin (Project 0420815101). 

\end{acknowledgement}



\bibliography{literature}

\providecommand{\latin}[1]{#1}
\makeatletter
\providecommand{\doi}
  {\begingroup\let\do\@makeother\dospecials
  \catcode`\{=1 \catcode`\}=2 \doi@aux}
\providecommand{\doi@aux}[1]{\endgroup\texttt{#1}}
\makeatother
\providecommand*\mcitethebibliography{\thebibliography}
\csname @ifundefined\endcsname{endmcitethebibliography}
  {\let\endmcitethebibliography\endthebibliography}{}
\begin{mcitethebibliography}{53}
\providecommand*\natexlab[1]{#1}
\providecommand*\mciteSetBstSublistMode[1]{}
\providecommand*\mciteSetBstMaxWidthForm[2]{}
\providecommand*\mciteBstWouldAddEndPuncttrue
  {\def\EndOfBibitem{\unskip.}}
\providecommand*\mciteBstWouldAddEndPunctfalse
  {\let\EndOfBibitem\relax}
\providecommand*\mciteSetBstMidEndSepPunct[3]{}
\providecommand*\mciteSetBstSublistLabelBeginEnd[3]{}
\providecommand*\EndOfBibitem{}
\mciteSetBstSublistMode{f}
\mciteSetBstMaxWidthForm{subitem}{(\alph{mcitesubitemcount})}
\mciteSetBstSublistLabelBeginEnd
  {\mcitemaxwidthsubitemform\space}
  {\relax}
  {\relax}

\bibitem[Clementi(2008)]{clementiCOSB2008}
Clementi,~C. Coarse-grained models of protein folding: toy models or predictive
  tools? \emph{Curr. Opin. Struct. Biol.} \textbf{2008}, \emph{18},
  10--15\relax
\mciteBstWouldAddEndPuncttrue
\mciteSetBstMidEndSepPunct{\mcitedefaultmidpunct}
{\mcitedefaultendpunct}{\mcitedefaultseppunct}\relax
\EndOfBibitem
\bibitem[Noid(2013)]{noid2013perspective}
Noid,~W.~G. Perspective: Coarse-grained models for biomolecular systems.
  \emph{J. Chem. Phys.} \textbf{2013}, \emph{139}, 09B201\_1\relax
\mciteBstWouldAddEndPuncttrue
\mciteSetBstMidEndSepPunct{\mcitedefaultmidpunct}
{\mcitedefaultendpunct}{\mcitedefaultseppunct}\relax
\EndOfBibitem
\bibitem[Saunders and Voth(2013)Saunders, and Voth]{saunders2013coarse}
Saunders,~M.~G.; Voth,~G.~A. Coarse-graining methods for computational biology.
  \emph{Annu. Rev. Biophys.} \textbf{2013}, \emph{42}, 73--93\relax
\mciteBstWouldAddEndPuncttrue
\mciteSetBstMidEndSepPunct{\mcitedefaultmidpunct}
{\mcitedefaultendpunct}{\mcitedefaultseppunct}\relax
\EndOfBibitem
\bibitem[Kmiecik \latin{et~al.}(2016)Kmiecik, Gront, Kolinski, Wieteska, Dawid,
  and Kolinski]{kmiecik2016coarse}
Kmiecik,~S.; Gront,~D.; Kolinski,~M.; Wieteska,~L.; Dawid,~A.~E.; Kolinski,~A.
  Coarse-grained protein models and their applications. \emph{Chem. Rev.}
  \textbf{2016}, \emph{116}, 7898--7936\relax
\mciteBstWouldAddEndPuncttrue
\mciteSetBstMidEndSepPunct{\mcitedefaultmidpunct}
{\mcitedefaultendpunct}{\mcitedefaultseppunct}\relax
\EndOfBibitem
\bibitem[Marrink \latin{et~al.}(2007)Marrink, Risselada, Yefimov, Tieleman, and
  De~Vries]{marrink2007martini}
Marrink,~S.~J.; Risselada,~H.~J.; Yefimov,~S.; Tieleman,~D.~P.; De~Vries,~A.~H.
  The MARTINI force field: coarse grained model for biomolecular simulations.
  \emph{J. Phys. Chem. B} \textbf{2007}, \emph{111}, 7812--7824\relax
\mciteBstWouldAddEndPuncttrue
\mciteSetBstMidEndSepPunct{\mcitedefaultmidpunct}
{\mcitedefaultendpunct}{\mcitedefaultseppunct}\relax
\EndOfBibitem
\bibitem[Davtyan \latin{et~al.}(2012)Davtyan, Schafer, Zheng, Clementi,
  Wolynes, and Papoian]{davtyan2012awsem}
Davtyan,~A.; Schafer,~N.~P.; Zheng,~W.; Clementi,~C.; Wolynes,~P.~G.;
  Papoian,~G.~A. AWSEM-MD: protein structure prediction using coarse-grained
  physical potentials and bioinformatically based local structure biasing.
  \emph{J. Phys. Chem. B} \textbf{2012}, \emph{116}, 8494--8503\relax
\mciteBstWouldAddEndPuncttrue
\mciteSetBstMidEndSepPunct{\mcitedefaultmidpunct}
{\mcitedefaultendpunct}{\mcitedefaultseppunct}\relax
\EndOfBibitem
\bibitem[Matysiak and Clementi(2006)Matysiak, and Clementi]{Matysiak2006}
Matysiak,~S.; Clementi,~C. Minimalist Protein Model as a Diagnostic Tool for
  Misfolding and Aggregation. \emph{J. Mol. Biol.} \textbf{2006}, \emph{363},
  297--308\relax
\mciteBstWouldAddEndPuncttrue
\mciteSetBstMidEndSepPunct{\mcitedefaultmidpunct}
{\mcitedefaultendpunct}{\mcitedefaultseppunct}\relax
\EndOfBibitem
\bibitem[Chen \latin{et~al.}(2018)Chen, Chen, Pinamonti, and
  Clementi]{Chen2018}
Chen,~J.; Chen,~J.; Pinamonti,~G.; Clementi,~C. Learning Effective Molecular
  Models from Experimental Observables. \emph{J. Chem. Theory Comput.}
  \textbf{2018}, \emph{14}, 3849--3858\relax
\mciteBstWouldAddEndPuncttrue
\mciteSetBstMidEndSepPunct{\mcitedefaultmidpunct}
{\mcitedefaultendpunct}{\mcitedefaultseppunct}\relax
\EndOfBibitem
\bibitem[Noid \latin{et~al.}(2008)Noid, Chu, Ayton, Krishna, Izvekov, Voth,
  Das, and Andersen]{noid2008multiscale}
Noid,~W.~G.; Chu,~J.-W.; Ayton,~G.~S.; Krishna,~V.; Izvekov,~S.; Voth,~G.~A.;
  Das,~A.; Andersen,~H.~C. The multiscale coarse-graining method. I. A rigorous
  bridge between atomistic and coarse-grained models. \emph{J. Chem. Phys.}
  \textbf{2008}, \emph{128}, 244114\relax
\mciteBstWouldAddEndPuncttrue
\mciteSetBstMidEndSepPunct{\mcitedefaultmidpunct}
{\mcitedefaultendpunct}{\mcitedefaultseppunct}\relax
\EndOfBibitem
\bibitem[Shell(2008)]{shell2008relative}
Shell,~M.~S. The relative entropy is fundamental to multiscale and inverse
  thermodynamic problems. \emph{J. Chem. Phys.} \textbf{2008}, \emph{129},
  144108\relax
\mciteBstWouldAddEndPuncttrue
\mciteSetBstMidEndSepPunct{\mcitedefaultmidpunct}
{\mcitedefaultendpunct}{\mcitedefaultseppunct}\relax
\EndOfBibitem
\bibitem[N\"{u}ske \latin{et~al.}(2019)N\"{u}ske, Boninsegna, and
  Clementi]{Nueske2019}
N\"{u}ske,~F.; Boninsegna,~L.; Clementi,~C. Coarse-graining molecular systems
  by spectral matching. \emph{J. Chem. Phys.} \textbf{2019}, \emph{151},
  044116\relax
\mciteBstWouldAddEndPuncttrue
\mciteSetBstMidEndSepPunct{\mcitedefaultmidpunct}
{\mcitedefaultendpunct}{\mcitedefaultseppunct}\relax
\EndOfBibitem
\bibitem[Mim \latin{et~al.}(2012)Mim, Cui, Gawronski-Salerno, Frost, Lyman,
  Voth, and Unger]{Mim2012}
Mim,~C.; Cui,~H.; Gawronski-Salerno,~J.~A.; Frost,~A.; Lyman,~E.; Voth,~G.~A.;
  Unger,~V.~M. Structural Basis of Membrane Bending by the N-{BAR} Protein
  Endophilin. \emph{Cell} \textbf{2012}, \emph{149}, 137--145\relax
\mciteBstWouldAddEndPuncttrue
\mciteSetBstMidEndSepPunct{\mcitedefaultmidpunct}
{\mcitedefaultendpunct}{\mcitedefaultseppunct}\relax
\EndOfBibitem
\bibitem[Chu and Voth(2005)Chu, and Voth]{Chu2005}
Chu,~J.-W.; Voth,~G.~A. Allostery of actin filaments: Molecular dynamics
  simulations and coarse-grained analysis. \emph{Proc. Natl. Acad. Sci. U.S.A}
  \textbf{2005}, \emph{102}, 13111--13116\relax
\mciteBstWouldAddEndPuncttrue
\mciteSetBstMidEndSepPunct{\mcitedefaultmidpunct}
{\mcitedefaultendpunct}{\mcitedefaultseppunct}\relax
\EndOfBibitem
\bibitem[Yu \latin{et~al.}(2021)Yu, Pak, He, Monje-Galvan, Casalino, Gaieb,
  Dommer, Amaro, and Voth]{Yu2021}
Yu,~A.; Pak,~A.~J.; He,~P.; Monje-Galvan,~V.; Casalino,~L.; Gaieb,~Z.;
  Dommer,~A.~C.; Amaro,~R.~E.; Voth,~G.~A. A multiscale coarse-grained model of
  the {SARS}-{CoV}-2 virion. \emph{Biophys. J.} \textbf{2021}, \emph{120},
  1097--1104\relax
\mciteBstWouldAddEndPuncttrue
\mciteSetBstMidEndSepPunct{\mcitedefaultmidpunct}
{\mcitedefaultendpunct}{\mcitedefaultseppunct}\relax
\EndOfBibitem
\bibitem[Wang \latin{et~al.}(2019)Wang, Olsson, Wehmeyer, P{\'e}rez, Charron,
  De~Fabritiis, No{\'e}, and Clementi]{wang2019machine}
Wang,~J.; Olsson,~S.; Wehmeyer,~C.; P{\'e}rez,~A.; Charron,~N.~E.;
  De~Fabritiis,~G.; No{\'e},~F.; Clementi,~C. Machine learning of
  coarse-grained molecular dynamics force fields. \emph{ACS Cent. Sci.}
  \textbf{2019}, \emph{5}, 755--767\relax
\mciteBstWouldAddEndPuncttrue
\mciteSetBstMidEndSepPunct{\mcitedefaultmidpunct}
{\mcitedefaultendpunct}{\mcitedefaultseppunct}\relax
\EndOfBibitem
\bibitem[Husic \latin{et~al.}(2020)Husic, Charron, Lemm, Wang, P{\'e}rez,
  Majewski, Kr{\"a}mer, Chen, Olsson, de~Fabritiis, \latin{et~al.}
  others]{husic2020coarse}
Husic,~B.~E.; Charron,~N.~E.; Lemm,~D.; Wang,~J.; P{\'e}rez,~A.; Majewski,~M.;
  Kr{\"a}mer,~A.; Chen,~Y.; Olsson,~S.; de~Fabritiis,~G., \latin{et~al.}
  Coarse graining molecular dynamics with graph neural networks. \emph{J. Chem.
  Phys.} \textbf{2020}, \emph{153}, 194101\relax
\mciteBstWouldAddEndPuncttrue
\mciteSetBstMidEndSepPunct{\mcitedefaultmidpunct}
{\mcitedefaultendpunct}{\mcitedefaultseppunct}\relax
\EndOfBibitem
\bibitem[Song and Kingma(2021)Song, and Kingma]{song2021train}
Song,~Y.; Kingma,~D.~P. How to train your energy-based models. \emph{arXiv
  preprint arXiv:2101.03288} \textbf{2021}, \relax
\mciteBstWouldAddEndPunctfalse
\mciteSetBstMidEndSepPunct{\mcitedefaultmidpunct}
{}{\mcitedefaultseppunct}\relax
\EndOfBibitem
\bibitem[Hinton(2002)]{hinton2002training}
Hinton,~G.~E. Training products of experts by minimizing contrastive
  divergence. \emph{Neural Comput.} \textbf{2002}, \emph{14}, 1771--1800\relax
\mciteBstWouldAddEndPuncttrue
\mciteSetBstMidEndSepPunct{\mcitedefaultmidpunct}
{\mcitedefaultendpunct}{\mcitedefaultseppunct}\relax
\EndOfBibitem
\bibitem[K\"{o}hler \latin{et~al.}(2023)K\"{o}hler, Chen, Kr\"{a}mer, Clementi,
  and No{\'{e}}]{kohler2022force}
K\"{o}hler,~J.; Chen,~Y.; Kr\"{a}mer,~A.; Clementi,~C.; No{\'{e}},~F.
  Flow-Matching: Efficient Coarse-Graining of Molecular Dynamics without
  Forces. \emph{J. Chem. Theory Comput.} \textbf{2023}, \relax
\mciteBstWouldAddEndPunctfalse
\mciteSetBstMidEndSepPunct{\mcitedefaultmidpunct}
{}{\mcitedefaultseppunct}\relax
\EndOfBibitem
\bibitem[Rezende and Mohamed(2015)Rezende, and Mohamed]{rezende2015variational}
Rezende,~D.; Mohamed,~S. Variational inference with normalizing flows. Int.
  Conf. Mach. Learn. 2015; pp 1530--1538\relax
\mciteBstWouldAddEndPuncttrue
\mciteSetBstMidEndSepPunct{\mcitedefaultmidpunct}
{\mcitedefaultendpunct}{\mcitedefaultseppunct}\relax
\EndOfBibitem
\bibitem[Papamakarios \latin{et~al.}(2021)Papamakarios, Nalisnick, Rezende,
  Mohamed, and Lakshminarayanan]{papamakarios2021normalizing}
Papamakarios,~G.; Nalisnick,~E.~T.; Rezende,~D.~J.; Mohamed,~S.;
  Lakshminarayanan,~B. Normalizing Flows for Probabilistic Modeling and
  Inference. \emph{JMLR} \textbf{2021}, \emph{22}, 1--64\relax
\mciteBstWouldAddEndPuncttrue
\mciteSetBstMidEndSepPunct{\mcitedefaultmidpunct}
{\mcitedefaultendpunct}{\mcitedefaultseppunct}\relax
\EndOfBibitem
\bibitem[Huang \latin{et~al.}(2020)Huang, Dinh, and
  Courville]{huang2020augmented}
Huang,~C.-W.; Dinh,~L.; Courville,~A. Augmented normalizing flows: Bridging the
  gap between generative flows and latent variable models. \emph{arXiv preprint
  arXiv:2002.07101} \textbf{2020}, \relax
\mciteBstWouldAddEndPunctfalse
\mciteSetBstMidEndSepPunct{\mcitedefaultmidpunct}
{}{\mcitedefaultseppunct}\relax
\EndOfBibitem
\bibitem[Chen \latin{et~al.}(2020)Chen, Lu, Chenli, Zhu, and
  Tian]{chen2020vflow}
Chen,~J.; Lu,~C.; Chenli,~B.; Zhu,~J.; Tian,~T. Vflow: More expressive
  generative flows with variational data augmentation. Int. Conf. Mach. Learn.
  2020; pp 1660--1669\relax
\mciteBstWouldAddEndPuncttrue
\mciteSetBstMidEndSepPunct{\mcitedefaultmidpunct}
{\mcitedefaultendpunct}{\mcitedefaultseppunct}\relax
\EndOfBibitem
\bibitem[Lindorff-Larsen \latin{et~al.}(2011)Lindorff-Larsen, Piana, Dror, and
  Shaw]{lindorff2011fast}
Lindorff-Larsen,~K.; Piana,~S.; Dror,~R.~O.; Shaw,~D.~E. How fast-folding
  proteins fold. \emph{Science} \textbf{2011}, \emph{334}, 517--520\relax
\mciteBstWouldAddEndPuncttrue
\mciteSetBstMidEndSepPunct{\mcitedefaultmidpunct}
{\mcitedefaultendpunct}{\mcitedefaultseppunct}\relax
\EndOfBibitem
\bibitem[Ho \latin{et~al.}(2020)Ho, Jain, and Abbeel]{ho2020denoising}
Ho,~J.; Jain,~A.; Abbeel,~P. Denoising diffusion probabilistic models.
  \emph{Adv. Neural Inf. Process. Syst.} \textbf{2020}, \emph{33},
  6840--6851\relax
\mciteBstWouldAddEndPuncttrue
\mciteSetBstMidEndSepPunct{\mcitedefaultmidpunct}
{\mcitedefaultendpunct}{\mcitedefaultseppunct}\relax
\EndOfBibitem
\bibitem[Sohl-Dickstein \latin{et~al.}(2015)Sohl-Dickstein, Weiss,
  Maheswaranathan, and Ganguli]{sohl2015deep}
Sohl-Dickstein,~J.; Weiss,~E.; Maheswaranathan,~N.; Ganguli,~S. Deep
  Unsupervised Learning using Nonequilibrium Thermodynamics. Int. Conf. Mach.
  Learn. 2015; pp 2256--2265\relax
\mciteBstWouldAddEndPuncttrue
\mciteSetBstMidEndSepPunct{\mcitedefaultmidpunct}
{\mcitedefaultendpunct}{\mcitedefaultseppunct}\relax
\EndOfBibitem
\bibitem[Wu \latin{et~al.}(2022)Wu, Yang, Berg, Zou, Lu, and
  Amini]{wu2022protein}
Wu,~K.~E.; Yang,~K.~K.; Berg,~R. v.~d.; Zou,~J.~Y.; Lu,~A.~X.; Amini,~A.~P.
  Protein structure generation via folding diffusion. \emph{arXiv preprint
  arXiv:2209.15611} \textbf{2022}, \relax
\mciteBstWouldAddEndPunctfalse
\mciteSetBstMidEndSepPunct{\mcitedefaultmidpunct}
{}{\mcitedefaultseppunct}\relax
\EndOfBibitem
\bibitem[Trippe \latin{et~al.}(2022)Trippe, Yim, Tischer, Broderick, Baker,
  Barzilay, and Jaakkola]{trippe2022diffusion}
Trippe,~B.~L.; Yim,~J.; Tischer,~D.; Broderick,~T.; Baker,~D.; Barzilay,~R.;
  Jaakkola,~T. Diffusion probabilistic modeling of protein backbones in 3D for
  the motif-scaffolding problem. \emph{arXiv preprint arXiv:2206.04119}
  \textbf{2022}, \relax
\mciteBstWouldAddEndPunctfalse
\mciteSetBstMidEndSepPunct{\mcitedefaultmidpunct}
{}{\mcitedefaultseppunct}\relax
\EndOfBibitem
\bibitem[Watson \latin{et~al.}(2022)Watson, Juergens, Bennett, Trippe, Yim,
  Eisenach, Ahern, Borst, Ragotte, Milles, \latin{et~al.}
  others]{watson2022broadly}
Watson,~J.~L.; Juergens,~D.; Bennett,~N.~R.; Trippe,~B.~L.; Yim,~J.;
  Eisenach,~H.~E.; Ahern,~W.; Borst,~A.~J.; Ragotte,~R.~J.; Milles,~L.~F.,
  \latin{et~al.}  Broadly applicable and accurate protein design by integrating
  structure prediction networks and diffusion generative models. \emph{bioRxiv}
  \textbf{2022}, \relax
\mciteBstWouldAddEndPunctfalse
\mciteSetBstMidEndSepPunct{\mcitedefaultmidpunct}
{}{\mcitedefaultseppunct}\relax
\EndOfBibitem
\bibitem[Igashov \latin{et~al.}(2022)Igashov, St{\"a}rk, Vignac, Satorras,
  Frossard, Welling, Bronstein, and Correia]{igashov2022equivariant}
Igashov,~I.; St{\"a}rk,~H.; Vignac,~C.; Satorras,~V.~G.; Frossard,~P.;
  Welling,~M.; Bronstein,~M.; Correia,~B. Equivariant 3d-conditional diffusion
  models for molecular linker design. \emph{arXiv preprint arXiv:2210.05274}
  \textbf{2022}, \relax
\mciteBstWouldAddEndPunctfalse
\mciteSetBstMidEndSepPunct{\mcitedefaultmidpunct}
{}{\mcitedefaultseppunct}\relax
\EndOfBibitem
\bibitem[Qiao \latin{et~al.}(2022)Qiao, Nie, Vahdat, Miller~III, and
  Anandkumar]{qiao2022dynamic}
Qiao,~Z.; Nie,~W.; Vahdat,~A.; Miller~III,~T.~F.; Anandkumar,~A.
  Dynamic-Backbone Protein-Ligand Structure Prediction with Multiscale
  Generative Diffusion Models. \emph{arXiv preprint arXiv:2209.15171}
  \textbf{2022}, \relax
\mciteBstWouldAddEndPunctfalse
\mciteSetBstMidEndSepPunct{\mcitedefaultmidpunct}
{}{\mcitedefaultseppunct}\relax
\EndOfBibitem
\bibitem[Hoogeboom \latin{et~al.}(2022)Hoogeboom, Satorras, Vignac, and
  Welling]{hoogeboom2022equivariant}
Hoogeboom,~E.; Satorras,~V.~G.; Vignac,~C.; Welling,~M. Equivariant diffusion
  for molecule generation in 3d. International conference on machine learning.
  2022; pp 8867--8887\relax
\mciteBstWouldAddEndPuncttrue
\mciteSetBstMidEndSepPunct{\mcitedefaultmidpunct}
{\mcitedefaultendpunct}{\mcitedefaultseppunct}\relax
\EndOfBibitem
\bibitem[Jing \latin{et~al.}(2022)Jing, Corso, Chang, Barzilay, and
  Jaakkola]{jing2022torsional}
Jing,~B.; Corso,~G.; Chang,~J.; Barzilay,~R.; Jaakkola,~T. Torsional Diffusion
  for Molecular Conformer Generation. \emph{arXiv preprint arXiv:2206.01729}
  \textbf{2022}, \relax
\mciteBstWouldAddEndPunctfalse
\mciteSetBstMidEndSepPunct{\mcitedefaultmidpunct}
{}{\mcitedefaultseppunct}\relax
\EndOfBibitem
\bibitem[Corso \latin{et~al.}(2022)Corso, St{\"a}rk, Jing, Barzilay, and
  Jaakkola]{corso2022diffdock}
Corso,~G.; St{\"a}rk,~H.; Jing,~B.; Barzilay,~R.; Jaakkola,~T. Diffdock:
  Diffusion steps, twists, and turns for molecular docking. \emph{arXiv
  preprint arXiv:2210.01776} \textbf{2022}, \relax
\mciteBstWouldAddEndPunctfalse
\mciteSetBstMidEndSepPunct{\mcitedefaultmidpunct}
{}{\mcitedefaultseppunct}\relax
\EndOfBibitem
\bibitem[Song \latin{et~al.}(2020)Song, Sohl-Dickstein, Kingma, Kumar, Ermon,
  and Poole]{song2020score}
Song,~Y.; Sohl-Dickstein,~J.; Kingma,~D.~P.; Kumar,~A.; Ermon,~S.; Poole,~B.
  Score-based generative modeling through stochastic differential equations.
  \emph{arXiv preprint arXiv:2011.13456} \textbf{2020}, \relax
\mciteBstWouldAddEndPunctfalse
\mciteSetBstMidEndSepPunct{\mcitedefaultmidpunct}
{}{\mcitedefaultseppunct}\relax
\EndOfBibitem
\bibitem[Ciccotti \latin{et~al.}(2005)Ciccotti, Kapral, and
  Vanden-Eijnden]{Ciccotti2005}
Ciccotti,~G.; Kapral,~R.; Vanden-Eijnden,~E. Blue Moon sampling, vectorial
  reaction coordinates, and unbiased constrained dynamics. \emph{ChemPhysChem}
  \textbf{2005}, \emph{6}, 1809--1814\relax
\mciteBstWouldAddEndPuncttrue
\mciteSetBstMidEndSepPunct{\mcitedefaultmidpunct}
{\mcitedefaultendpunct}{\mcitedefaultseppunct}\relax
\EndOfBibitem
\bibitem[Thaler \latin{et~al.}(2022)Thaler, Stupp, and
  Zavadlav]{thaler2022deep}
Thaler,~S.; Stupp,~M.; Zavadlav,~J. Deep Coarse-grained Potentials via Relative
  Entropy Minimization. \emph{arXiv preprint arXiv:2208.10330} \textbf{2022},
  \relax
\mciteBstWouldAddEndPunctfalse
\mciteSetBstMidEndSepPunct{\mcitedefaultmidpunct}
{}{\mcitedefaultseppunct}\relax
\EndOfBibitem
\bibitem[Dinh \latin{et~al.}(2014)Dinh, Krueger, and Bengio]{dinh2014nice}
Dinh,~L.; Krueger,~D.; Bengio,~Y. Nice: Non-linear independent components
  estimation. \emph{arXiv preprint arXiv:1410.8516} \textbf{2014}, \relax
\mciteBstWouldAddEndPunctfalse
\mciteSetBstMidEndSepPunct{\mcitedefaultmidpunct}
{}{\mcitedefaultseppunct}\relax
\EndOfBibitem
\bibitem[Dinh \latin{et~al.}(2016)Dinh, Sohl-Dickstein, and
  Bengio]{dinh2016density}
Dinh,~L.; Sohl-Dickstein,~J.; Bengio,~S. Density estimation using real nvp.
  \emph{arXiv preprint arXiv:1605.08803} \textbf{2016}, \relax
\mciteBstWouldAddEndPunctfalse
\mciteSetBstMidEndSepPunct{\mcitedefaultmidpunct}
{}{\mcitedefaultseppunct}\relax
\EndOfBibitem
\bibitem[Vincent(2011)]{vincent2011connection}
Vincent,~P. A connection between score matching and denoising autoencoders.
  \emph{Neural computation} \textbf{2011}, \emph{23}, 1661--1674\relax
\mciteBstWouldAddEndPuncttrue
\mciteSetBstMidEndSepPunct{\mcitedefaultmidpunct}
{\mcitedefaultendpunct}{\mcitedefaultseppunct}\relax
\EndOfBibitem
\bibitem[Zaidi \latin{et~al.}(2022)Zaidi, Schaarschmidt, Martens, Kim, Teh,
  Sanchez-Gonzalez, Battaglia, Pascanu, and Godwin]{zaidi2022pre}
Zaidi,~S.; Schaarschmidt,~M.; Martens,~J.; Kim,~H.; Teh,~Y.~W.;
  Sanchez-Gonzalez,~A.; Battaglia,~P.; Pascanu,~R.; Godwin,~J. Pre-training via
  Denoising for Molecular Property Prediction. \emph{arXiv preprint
  arXiv:2206.00133} \textbf{2022}, \relax
\mciteBstWouldAddEndPunctfalse
\mciteSetBstMidEndSepPunct{\mcitedefaultmidpunct}
{}{\mcitedefaultseppunct}\relax
\EndOfBibitem
\bibitem[Xie \latin{et~al.}(2022)Xie, Fu, Ganea, Barzilay, and
  Jaakkola]{xie2021crystal}
Xie,~T.; Fu,~X.; Ganea,~O.-E.; Barzilay,~R.; Jaakkola,~T.~S. Crystal Diffusion
  Variational Autoencoder for Periodic Material Generation. Int. Conf. Learn.
  Represent. 2022\relax
\mciteBstWouldAddEndPuncttrue
\mciteSetBstMidEndSepPunct{\mcitedefaultmidpunct}
{\mcitedefaultendpunct}{\mcitedefaultseppunct}\relax
\EndOfBibitem
\bibitem[Salimans and Ho(2021)Salimans, and Ho]{salimans2021should}
Salimans,~T.; Ho,~J. Should EBMs model the energy or the score? Energy Based
  Models Workshop-ICLR. 2021\relax
\mciteBstWouldAddEndPuncttrue
\mciteSetBstMidEndSepPunct{\mcitedefaultmidpunct}
{\mcitedefaultendpunct}{\mcitedefaultseppunct}\relax
\EndOfBibitem
\bibitem[Satorras \latin{et~al.}(2021)Satorras, Hoogeboom, and
  Welling]{satorras2021n}
Satorras,~V.~G.; Hoogeboom,~E.; Welling,~M. E (n) equivariant graph neural
  networks. Int. Conf. Mach. Learn. 2021; pp 9323--9332\relax
\mciteBstWouldAddEndPuncttrue
\mciteSetBstMidEndSepPunct{\mcitedefaultmidpunct}
{\mcitedefaultendpunct}{\mcitedefaultseppunct}\relax
\EndOfBibitem
\bibitem[Gruver \latin{et~al.}(2022)Gruver, Finzi, Goldblum, and
  Wilson]{gruver2022lie}
Gruver,~N.; Finzi,~M.; Goldblum,~M.; Wilson,~A.~G. The Lie Derivative for
  Measuring Learned Equivariance. \emph{arXiv preprint arXiv:2210.02984}
  \textbf{2022}, \relax
\mciteBstWouldAddEndPunctfalse
\mciteSetBstMidEndSepPunct{\mcitedefaultmidpunct}
{}{\mcitedefaultseppunct}\relax
\EndOfBibitem
\bibitem[Naritomi and Fuchigami(2011)Naritomi, and Fuchigami]{naritomi2011slow}
Naritomi,~Y.; Fuchigami,~S. Slow dynamics in protein fluctuations revealed by
  time-structure based independent component analysis: the case of domain
  motions. \emph{J. Chem. Phys.} \textbf{2011}, \emph{134}, 02B617\relax
\mciteBstWouldAddEndPuncttrue
\mciteSetBstMidEndSepPunct{\mcitedefaultmidpunct}
{\mcitedefaultendpunct}{\mcitedefaultseppunct}\relax
\EndOfBibitem
\bibitem[P{\'e}rez-Hern{\'a}ndez \latin{et~al.}(2013)P{\'e}rez-Hern{\'a}ndez,
  Paul, Giorgino, De~Fabritiis, and No{\'e}]{perez2013identification}
P{\'e}rez-Hern{\'a}ndez,~G.; Paul,~F.; Giorgino,~T.; De~Fabritiis,~G.;
  No{\'e},~F. Identification of slow molecular order parameters for Markov
  model construction. \emph{J. Chem. Phys.} \textbf{2013}, \emph{139},
  07B604\_1\relax
\mciteBstWouldAddEndPuncttrue
\mciteSetBstMidEndSepPunct{\mcitedefaultmidpunct}
{\mcitedefaultendpunct}{\mcitedefaultseppunct}\relax
\EndOfBibitem
\bibitem[Schwantes and Pande(2013)Schwantes, and
  Pande]{schwantes2013improvements}
Schwantes,~C.~R.; Pande,~V.~S. Improvements in Markov state model construction
  reveal many non-native interactions in the folding of NTL9. \emph{J. Chem.
  Theory Comput.} \textbf{2013}, \emph{9}, 2000--2009\relax
\mciteBstWouldAddEndPuncttrue
\mciteSetBstMidEndSepPunct{\mcitedefaultmidpunct}
{\mcitedefaultendpunct}{\mcitedefaultseppunct}\relax
\EndOfBibitem
\bibitem[Hoffmann \latin{et~al.}(2021)Hoffmann, Scherer, Hempel, Mardt,
  de~Silva, Husic, Klus, Wu, Kutz, Brunton, \latin{et~al.}
  others]{hoffmann2021deeptime}
Hoffmann,~M.; Scherer,~M.; Hempel,~T.; Mardt,~A.; de~Silva,~B.; Husic,~B.~E.;
  Klus,~S.; Wu,~H.; Kutz,~N.; Brunton,~S.~L., \latin{et~al.}  Deeptime: a
  Python library for machine learning dynamical models from time series data.
  \emph{Mach. Learn.: Sci. Technol.} \textbf{2021}, \emph{3}, 015009\relax
\mciteBstWouldAddEndPuncttrue
\mciteSetBstMidEndSepPunct{\mcitedefaultmidpunct}
{\mcitedefaultendpunct}{\mcitedefaultseppunct}\relax
\EndOfBibitem
\bibitem[Prinz \latin{et~al.}(2011)Prinz, Wu, Sarich, Keller, Senne, Held,
  Chodera, Sch\"{u}tte, and No{\'{e}}]{Prinz2011}
Prinz,~J.-H.; Wu,~H.; Sarich,~M.; Keller,~B.; Senne,~M.; Held,~M.;
  Chodera,~J.~D.; Sch\"{u}tte,~C.; No{\'{e}},~F. Markov models of molecular
  kinetics: Generation and validation. \emph{J. Chem. Phys.} \textbf{2011},
  \emph{134}, 174105\relax
\mciteBstWouldAddEndPuncttrue
\mciteSetBstMidEndSepPunct{\mcitedefaultmidpunct}
{\mcitedefaultendpunct}{\mcitedefaultseppunct}\relax
\EndOfBibitem
\bibitem[Jin \latin{et~al.}(2022)Jin, Schweizer, and
  Voth]{jin2022understanding}
Jin,~J.; Schweizer,~K.~S.; Voth,~G.~A. Understanding dynamics in coarse-grained
  models: I. Universal excess entropy scaling relationship. \emph{J. Chem.
  Phys.} \textbf{2022}, \relax
\mciteBstWouldAddEndPunctfalse
\mciteSetBstMidEndSepPunct{\mcitedefaultmidpunct}
{}{\mcitedefaultseppunct}\relax
\EndOfBibitem
\bibitem[{Schrödinger}(version 2.5.2)]{PyMOL}
{Schrödinger}, \emph{The {PyMOL} Molecular Graphics System}; version
  2.5.2\relax
\mciteBstWouldAddEndPuncttrue
\mciteSetBstMidEndSepPunct{\mcitedefaultmidpunct}
{\mcitedefaultendpunct}{\mcitedefaultseppunct}\relax
\EndOfBibitem
\end{mcitethebibliography}


\renewcommand{\thefigure}{S\arabic{figure}}
\renewcommand{\thetable}{S\arabic{table}}
\renewcommand{\theequation}{S\arabic{equation}}
\renewcommand{\thesection}{\Alph{section}}
\setcounter{figure}{0} 
\setcounter{table}{0} 
\setcounter{equation}{0} 
\setcounter{section}{0} 

\newpage
\section{Derivations}

\subsection{Relation between Score Function $s_{\theta}(\cg_i, i)$ and Noise Predicting Network $\epsilon_\theta(\cg_i, i)$} \label{ap:relation_betweens_score_and_noise}

In this section, we show that Eq. 3 and 4 in the main paper are equal up to a reweighting of the summands by setting: $$s_{\theta}(\cg_i, i) = -\frac{\epsilon(\cg_i,i)}{\sqrt{1-\bar\alpha_i}}.$$


We start from Eq. 4:
\begin{align*}
  & \E_{\cg_i, \cg_0}\left[ \norm{s_{\theta}(\cg_i, i) - \nabla_{\cg_i} \log q(\cg_i \mid  \cg_0)}^2\right] & \\
  = &  \E_{\cg_i, \cg_0} \left[\norm{s_{\theta}(\cg_i, i) + \frac{\cg_i - \sqrt{\bar\alpha_i}\cg_0}{1-\bar\alpha_i}}^2 \right]  & \text{Substitute $q(\cg_i\mid\cg_0)$ = $\gN(\sqrt{\bar{\alpha}_i}\cg_0, (1-\bar{\alpha}_i)\mI) $} \\
=  &  \E_{\epsilon, \cg_0} \left[\!\vphantom{\norm{\frac{\sqrt{\bar{\alpha}_i}}{\bar\alpha_i}}^2}\left\lVert \vphantom{\frac{\sqrt{\bar{\alpha}_i}}{\bar\alpha_i}} s_{\theta}(\sqrt{\bar{\alpha}_i}\cg_0 +  \sqrt{1  -  \bar{\alpha}_i}\epsilon, i) \ + \right. \right. &\\
& \left. \left. \qquad \frac{\sqrt{\bar{\alpha}_i}\cg_0 +  \sqrt{1  -  \bar{\alpha}_i}\epsilon - \sqrt{\bar\alpha_i}\cg_0}{1-\bar\alpha_i} \right\rVert^2 \right] & \text{Reparameterize $\cg_i = \sqrt{\bar{\alpha}_i}\cg_0 +  \sqrt{1  -  \bar{\alpha}_i}\epsilon $} \\ 
=  &  \E_{\epsilon, \cg_0} \left[\norm{s_{\theta}(\sqrt{\bar{\alpha}_i}\cg_0 +  \sqrt{1  -  \bar{\alpha}_i}\epsilon, i) + \frac{\epsilon }{\sqrt{1-\bar\alpha_i}}}^2 \right] & \\
=  &  \E_{\epsilon, \cg_0} \left[\norm{ - \frac{\epsilon_\theta(\sqrt{\bar{\alpha}_i}\cg_0 +  \sqrt{1  -  \bar{\alpha}_i}\epsilon, i)}{\sqrt{1-\bar{\alpha}_i}} + \frac{\epsilon }{\sqrt{1-\bar\alpha_i}}}^2 \right] & \text{Substitute $s_{\theta}(\cdot, i) = - \frac{\epsilon_\theta(\cdot, i)}{\sqrt{1-\bar{\alpha}_i}}$}  \\
=  &  \frac{1}{1-\bar\alpha_i}\E_{\epsilon, \cg_0} \left[\norm{\epsilon - \epsilon_\theta(\sqrt{\bar{\alpha}_i}\cg_0 +  \sqrt{1  -  \bar{\alpha}_i}, i) }^2 \right]. & \text{Reorganize}
\end{align*}

That is, the noise-prediction loss of Eq. 3 can be seen as a reweighted denoising score matching loss, and vice versa.

\subsection{Connecting Denoising Diffusion and the Brownian Dynamics}\label{ap:connecting_ddpm_brownian}

Recall from Section 3 that a denoising-diffusion model consists of an iterative diffusion process $q(\cg_i \mid \cg_{i-1}) = \gN(\cg_i ; \sqrt{1-\beta_i} \cg_{i-1}, \beta_i \mI)$ and a denoising process $ p(\cg_{i-1} \mid \cg_{i}) = \gN(\cg_{i-1} ; \mu_\theta(\cg_{i}, i), \sigma_i^2 \mI)$ of $L$ steps, where $\beta_i$ and $\sigma_i^2$ are the variance coefficients of the diffusion and denoising steps, respectively, and $i \in [1, \dots, L]$. 
In this section, we show that iteratively performing a diffusion step followed by a denoising step at a low-noise level (e.g. $i=1$) approximates the Brownian dynamics of a molecular system. 
We demonstrate that by tuning the first diffusion noise level $\beta_1$, we effectively modify the step size of a Brownian dynamics simulation, as the two can be related by $ \beta_1 = 1-\bar\alpha_1 = \Delta t \frac{\temp}{M\gamma}$. We start by re-parameterizing the diffusion step $\cg_1 \sim q(\cg_1 \mid  \cg_0)$ as:

\begin{equation} \label{eq:diffusion_1}
        \cg_1 = \sqrt{1-\beta_1}\cg_0 + \sqrt{\beta_1}\mathbf{w}_{a},
\end{equation}

where $\mathbf{w}_{a} \sim \gN(0, \mathbf{I})$. 
Similarly, we re-parameterize $ \cg_{0} \sim p(\cg_{0} \mid \cg_{1}) = \gN(\cg_{0} ; \mu_\theta(\cg_{1}, 1), \sigma_1^2 \mI)$ and rewrite $\mu_\theta$ using the noise-predicting network $\epsilon_\theta$ from Section 3 as:

\begin{equation} \label{eq:app:denoise_1}
    \cg_0 = \frac{1}{\sqrt{1-\sigma_1}}\big(\cg_1 - \sqrt{\sigma_1}\mathbf{\epsilon}_\theta(\cg_1, 1)\big)  + \sqrt{\sigma_1}\mathbf{w}_b.
\end{equation}

Following \citet{ho2020denoising}, we set $ \sigma_i = \beta_i $. 
We now introduce a superscript index $(t)$ to denote the sequence of states that unfold the diffusion-denoising process through time.
Combining this notation with both Eq. \ref{eq:diffusion_1} and \ref{eq:app:denoise_1} we obtain the following recursive update:

\begin{align*} 
        \small\cg_1^{(t+1)} 
        &= \sqrt{1-\beta_1}
        \left[\frac{1}{\sqrt{1-\beta_1}}\big(\cg_1^{(t)} - \sqrt{\beta_1}\mathbf{\epsilon}_\theta(\cg_1^{(t)}, 1)\big)  + \right. & \\
        &\left. \vphantom{\frac{1}{\sqrt{1-\beta_1}}} \quad \; \sqrt{\beta_1}\mathbf{w}_{b} \right] + \sqrt{\beta_1}\mathbf{w}_{a} & \text{Combine  Eq. \ref{eq:diffusion_1} and \ref{eq:app:denoise_1}}  \\
        &= 
       \cg_1^{(t)} - \sqrt{\beta_1}\mathbf{\epsilon}_\theta(\cg_1^{(t)}, 1)  + \sqrt{1-\beta_1}\sqrt{\beta_1}\mathbf{w}_{b} + &\\
       &\quad \; \sqrt{\beta_1}\mathbf{w}_{a} & \text{Re-arrange terms}  \\
                &\overset{d}{=} 
       \cg_1^{(t)} - \sqrt{\beta_1}\mathbf{\epsilon}_\theta(\cg_1^{(t)}, 1) + \sqrt{2\beta_1 - \beta_1 ^2}\mathbf{w}_{c} & \text{Sum of the squares of the standard devs.}  \\
        &\approx 
       \cg_1^{(t)} - \sqrt{\beta_1}\mathbf{\epsilon}_\theta(\cg_1^{(t)}, 1) + \sqrt{2\beta_1}\mathbf{w}_{c} & \text{Taylor of } 2\beta_1 - \beta_1^2 \text{ at $(\beta_1 = 0) \rightarrow 2\beta_1$}  \\
        &= 
      \cg_1^{(t)} - \beta_1\frac{\mathbf{\epsilon}_\theta(\cg_1^{(t)}, 1)}{\sqrt{\beta_1}} + \sqrt{2\beta_1}\mathbf{w}_{c}, &  \text{Re-arrange} \\
\end{align*}

where $\mathbf{w}_c \sim \gN(0, \mathbf{I})$ and $\overset{d}{=}$ denotes equal in distribution. To make the relation to Brownian dynamics more explicit, we can discretize the overdamped limit of the Langevin dynamics from Eq. 6 using the Euler-Maruyama scheme:
\begin{equation}
    \mathrm \cg^{(t+1)}  = \mathrm \cg^{(t)} -\mathrm \Delta t \frac{\nabla_\cg V(\cg) }{\gamma M}   + \sqrt{\mathrm \Delta t\frac{ 2  \temp }{\gamma M}}\rmw,
\end{equation}   
where $\rmw \sim \mathcal{N}(0, \mathbf{I})$ and $\Delta t$ is the simulation time step. 
By comparing the two equations, we see that the denoising-diffusion process corresponds to simulating a Brownian dynamics of a force field $\nabla_\cg V(\cg)=\frac{\temp}{\sqrt{\beta_1}}\mathbf{\epsilon}_{\theta}(\cg, i) =  \frac{\temp}{\sqrt{1-\bar\alpha_1}}\mathbf{\epsilon}_{\theta}(\cg, i)$ 
and an implicit step size $\Delta t= \frac{M\gamma}{\temp}\beta_1$.
Note that the force field is the same as the denoising force field (Eq. 5).
Furthermore, the variance of the denoising-diffusion process $\beta_1=1-\bar{\alpha}_1$ is proportional to the simulation time step $\Delta t$.



\clearpage

\section{Additional Experiments}

\subsection{Ablation on Conservative Forces} \label{sec:ap_conservative_ablation}
Designing our neural network to approximate a conservative force is necessary to satisfy the following property in physics: \textit{``A conservative force is a force with the property that the total work done in moving a particle between two points is independent of the path taken.''}. This property can be satisfied by computing the force as the gradient of a scalar with respect to the input coordinates. In this section, we empirically show the benefits of using a conservative network when simulating dynamics (for \textit{DFF sim.}). In contrast, the benefits for i.i.d. generation (\textit{DFF i.i.d.}) are smaller.

We build a non-conservative network by replacing the last linear layer and average pooling operation of the conservative network, such that the network $\hat{\mathrm{nn}}_\theta: \mathbb{R}^{n\times 3} \rightarrow \mathbb{R}^{n \times 3}$ outputs a three-dimensional vector for each node (instead of a scalar). The non-conservative noise prediction network is then defined as $\epsilon_\theta = \hat {\mathrm{nn}}_\theta$. 

Next, we compare the performance of both variants on the proteins Chignolin and Trp-cage using the same experimental settings as in \cref{sec:fastfolders}.

\begin{figure} [ht]
    \centering
    \includegraphics[width=0.9\textwidth]{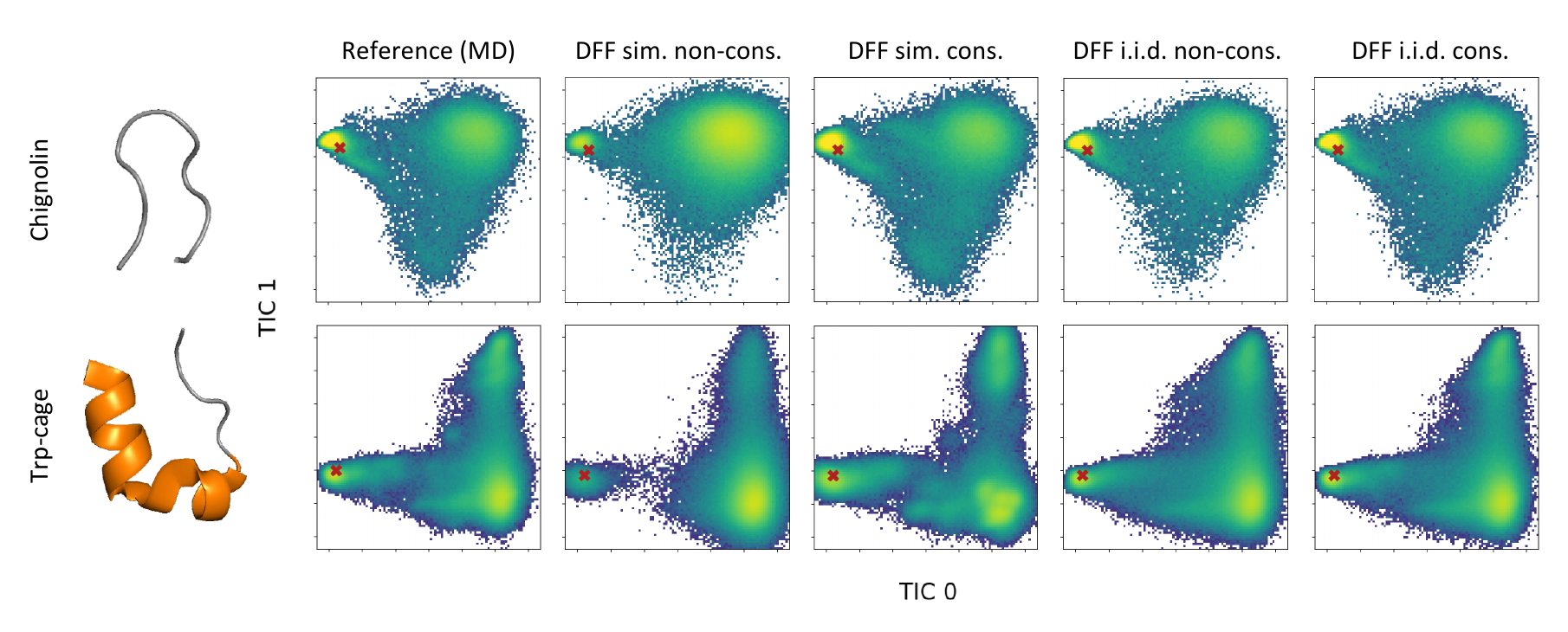}
    \caption{Ablation on TICA plots for chignolin and trp-cage fast folders comparing conservative vs non-conservative neural networks for simulated and i.i.d. data with our DFF model.}
    \label{fig:cons_vs_nocons}
\end{figure}


\begin{table}[ht]
\setlength{\tabcolsep}{4pt}
\centering
\caption{Ablation study on conservative versus non-conservative variants in Chignolin and Trp-cage. We report the JS divergence between TICA distributions and pairwise distance distributions.}
\label{tab:ablation_conservativeness}
\vspace{4pt}
\begin{tabular}{lccccc}
\toprule
 &  & \multicolumn{2}{c}{Chignolin} & \multicolumn{2}{c}{Trp-cage} \\
 &  & TIC JS & PWD JS & TIC JS & PWD JS \\ \midrule
Reference &  & .0057 & .0002 & .0026 & .0002 \\ \midrule
DFF (non-conservative) & \multirow{2}{*}{i.i.d.} & .0104 & .0008 & .0120 & .0030 \\
DFF &  & .0096 & .0005 & .0052 & .0007 \\ \midrule
DFF (non-conservative) & \multirow{2}{*}{sim.} & .3216 & .2147 & .0879 & .0399 \\
DFF &  & .0335 & .0067 & .0518 & .0403 \\ \bottomrule
\end{tabular}%
\end{table}

As illustrated in \cref{fig:cons_vs_nocons}, and numerically shown in \cref{tab:ablation_conservativeness}, the conservative variant outperforms the non-conservative case in most metrics for Chignolin and Trp-cage. Notice this difference is very significant in Chignolin simulated data where the TIC JS performance differs by an order of magnitude and even more for PWD JS. In Chignolin, we also tried using a more powerful non-conservative network, and while the non-conservative dynamics would improve for deeper networks, the performance would still be quite far from the conservative DFF sim. metric reported in the current experiment. On the other hand, the performance gap between conservative and non-conservative force fields in i.i.d. generation was reduced when we increased the size of the network.

\subsection{Ablation on Number of Hidden Features} \label{sec:ap_nf_ablation}
In \cref{sec:fastfolders}, we mention that the expressivity of the neural network (e.g. number of hidden features) has a notable influence on the quality of the simulated dynamics. More specifically, while i.i.d. generation tends to perform better as we increase the expressivity of our network, dynamics simulations may become worse. We found that properly choosing the number of features in our network plays an important role in the quality of the generated samples. Here, we show that the sweet spot in the bias/variance trade-off is significantly more sensitive when using the force field to simulate dynamics than when doing i.i.d. generation with the same network. We report the performance of DFF (i.i.d. and sim.) for different number of hidden features $\{32, 64, 128,256 \}$. All other training settings are the same as described for Chignolin in the fast-folding protein experiment in \cref{ap:fastfolders_implementation_details}.

\begin{figure} [ht]
    \centering
    \includegraphics[width=0.9\textwidth]{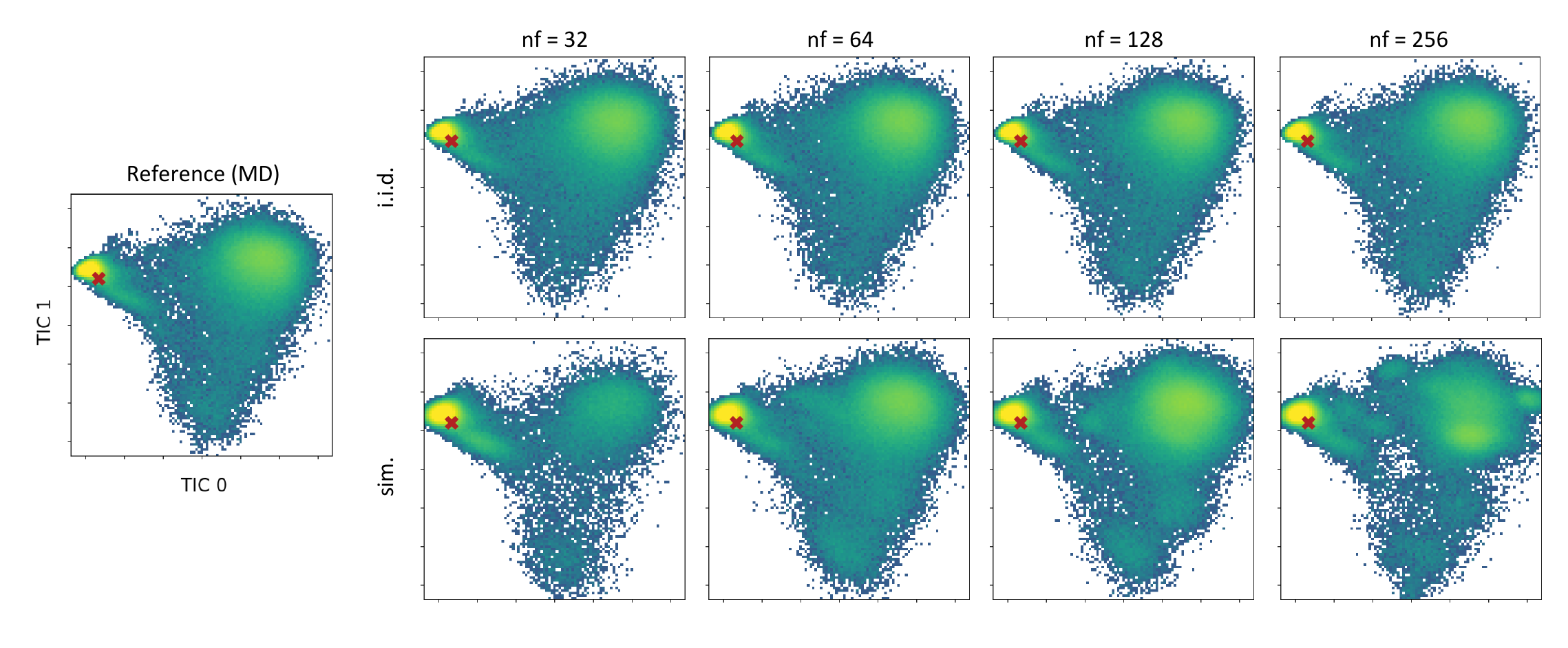}
    \caption{TICA plots for the ablation on the number of features per layer for Chignolin.}
    \label{fig:ablation_nf}
    \vspace{-10pt}
\end{figure}

\cref{fig:ablation_nf} and \cref{tab:ablation_nf} show that while the performance of i.i.d. generation improves as we increase the number of hidden features, the performance of simulated dynamics peaks at 64 features and deteriorates as we increase the network size. This implies that dynamics simulation is more sensitive to the number of features compared to i.i.d. sampling. The number of training iterations was set to 1M in all models and the validation curves on noise-prediction loss did not show symptoms of over-fitting.

\begin{table}[ht]
\centering
\setlength{\tabcolsep}{4pt}
\caption{TIC JS and PWD JS w.r.t the test partition with varying numbers of features. The model used to compute all metrics is our proposed DFF in both i.i.d. sampling and dynamics simulation.}
\label{tab:ablation_nf}
\vspace{4pt}
\begin{tabular}{lccc}
\toprule
 & Number of features & \multicolumn{2}{c}{Chignolin} \\
 &  & TIC JS & PWD JS \\ \midrule
Reference &  & .0057 & .0002 \\ \midrule
\multirow{4}{*}{i.i.d.} & 32 & .0105 & .0010 \\
 & 64 & .0096 & .0005 \\
 & 128 & \textbf{.0091} & \textbf{.0004} \\
 & 256 & \textbf{.0091} & \textbf{.0004} \\ \midrule
\multirow{4}{*}{sim.} & 32 & .0692 & .0318 \\
 & 64 & \textbf{.0335} & \textbf{.0067} \\
 & 128 & .0434 & .0179 \\
 & 256 & .0503 & .0263 \\ \bottomrule
\end{tabular}%
\end{table}

The proposed method (DFF sim.) has shown excellent performance in systems such as Alanine, Chignolin (nf=64) and Villin compared to previous ML coarse grained methods. Despite its good performance, we believe improving the robustness of the model by introducing new inductive biases or regularization techniques can be a promising direction in order to leverage higher-capacity neural networks without compromising the performance in simulation. Such increased robustness could also improve the performance for systems with a stronger dependence on non-local contacts and therefore a more challenging free energy distribution, like proteins with a higher $\beta$-sheet content (e.g. Bba, Protein G).

\subsection{Ablation on noise level.}
In this section, we assess the performance across various noise levels. As mentioned in Section 3.1, the force field derived from the noise prediction network $\mathbf{\epsilon}_{\theta^*}(\cg, i)$ approximates the data distribution for low noise levels $i$.
However, there are two types of errors in the approximation of the approximate force. The first type of error comes from approximating $q(\cg)$ as $q(\cg_i)$, where ideally $q(\cg_i)$ should approximate $q(\cg)$ for small noise levels. The second type of error comes from the learned model not being able to perfectly fit $q(\cg_i)$. Our network $\epsilon_\theta(\cg_i, i)$ is trained in the context of a limited number of data points, where larger noise values may serve as an inductive bias to facilitate interpolation between data points. Consequently, the optimal noise level $i$ is in practice determined by a balance between the amount of noise and the quantity of available data (as well as the network capacity). A smaller dataset may therefore require a larger noise value.

\begin{figure} [h]
    \centering
    \includegraphics[width=1.0\textwidth]{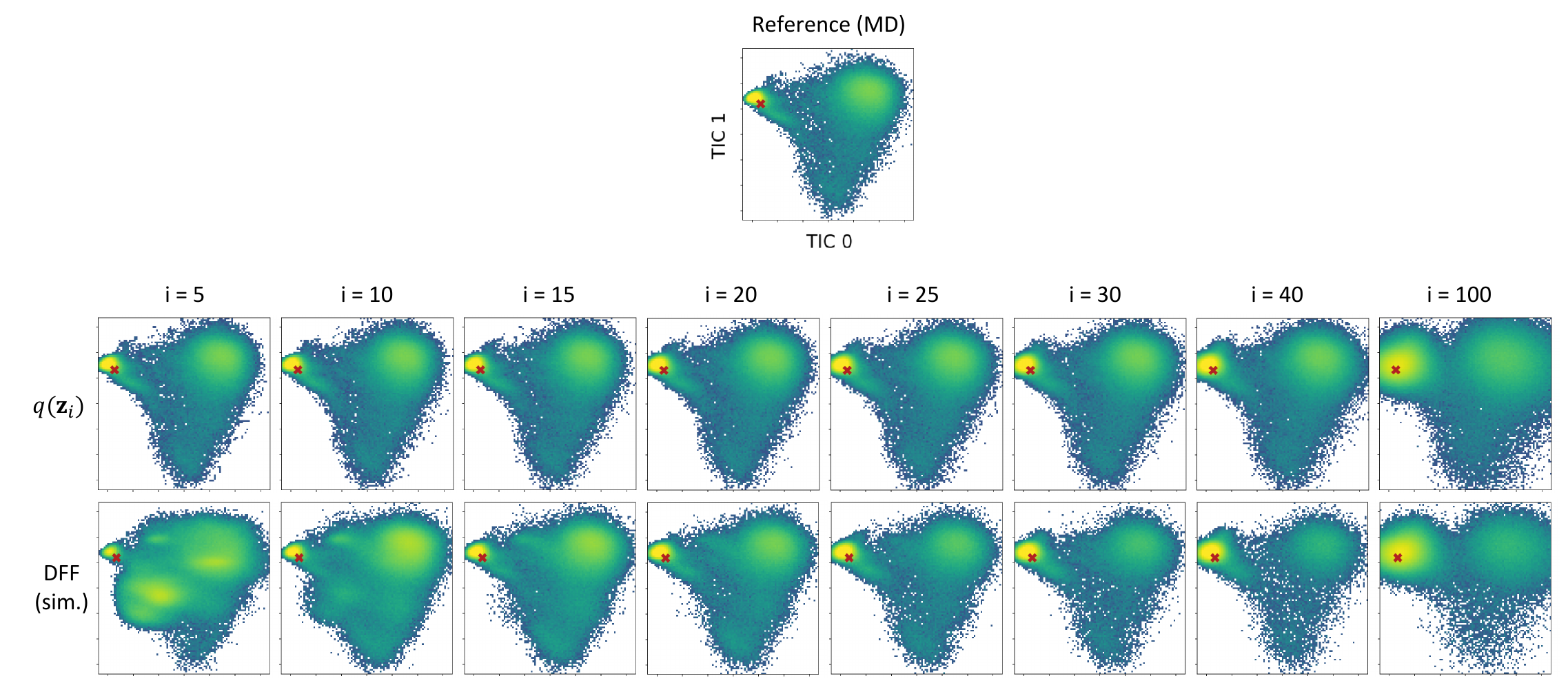}
    \caption{TICA plots for the ablation on different noise levels for Chignolin.}
    \label{fig:noise_levels}
    \vspace{-10pt}
\end{figure}

To demonstrate this behavior, we present TICA plots for Chignolin with different noise levels $i$ in \cref{fig:noise_levels}. These plots display two aspects: 1) the true distribution $q(\cg_i)$, obtained by convolving the data distribution $q(\cg_0)$ with a Gaussian kernel $ q(\cg_i | \cg_0)q(\cg_0)$, and 2) the DFF simulation, achieved by running Langevin Dynamics with the extracted force field. As can be observed in \cref{fig:noise_levels}, $q(\cg_i)$ more closely resembles the reference distribution for smaller noise levels; however, in practice, when running Langevin Dynamics with the learned force field (DFF sim.), we find a trade-off in the chosen noise level $i$.

\subsection{Rotation Equivariance through Data Augmentation} \label{sec:ap_equivariance_ablation}
In this section, we experimentally confirm that using data augmentation for training the neural network $\epsilon_\theta$ to be approximately equivariant to rotations is sufficient in practice. Recall that a function $f$ is equivariant to rotations if rotating its input results in an equivalent rotation of its output, i.e. $\mathbf{R}f(\cg) = f(\mathbf{R}\cg)$, where $\mathbf{R}$ is a rotation matrix. To analyze the degree of equivariance of our network $\epsilon_\theta$, we compute the relative squared error between $\epsilon_\theta(\cg)$ and $\mathbf{R}^{-1}\epsilon_\theta(\mathbf{R}\cg)$ for samples $\cg \sim p_{\text{val}}(\cg)$ drawn from the validation partition. 
In \cref{tab:equivariance}, we show the errors of the networks $\epsilon_\theta(\cdot,i)$ trained on different proteins. Specifically, we evaluate the non-equivariance at the noise level $i$ at which we extract the force field to simulate Langevin dynamics. These noise levels are $\{20, 15, 5, 5, 5\}$ for Chignolin, Trp-cage, Bba, Villin and Protein G, respectively. Since the relative squared error is in the order of $<10^{-6}$ for all proteins, we conclude that applying data augmentation is sufficient for achieving rotation equivariance for most practical use cases.

\begin{table}[h]
\centering
\setlength{\tabcolsep}{4pt}
\caption{Relative squared error between $\epsilon_\theta(\cg)$ and $\mathbf{R}^{-1}\epsilon_\theta(\mathbf{R}\cg)$ across different fast-folding proteins.}
\label{tab:equivariance}
\vspace{4pt}
\begin{tabular}{@{}lrrrrr@{}}
\toprule
 & \multicolumn{1}{c}{Chignolin} & \multicolumn{1}{c}{Trp-cage} & \multicolumn{1}{c}{Bba} & \multicolumn{1}{c}{Villin} & \multicolumn{1}{c}{Protein G} \\ \midrule
Relative squared error (mean) & $7.8 \cdot 10^{-7}$ & $4.9 \cdot 10^{-7}$ & $2.8\cdot 10^{-7}$ & $1.9\cdot 10^{-7}$ & $1.9\cdot 10^{-7}$ \\
Relative squared error (stdev) & $8.7 \cdot 10^{-7}$ & $6.0 \cdot 10^{-7}$ & $1.6\cdot 10^{-7}$ & $6.5\cdot 10^{-8}$ & $7.0\cdot 10^{-8}$ \\ \bottomrule
\end{tabular}%
\end{table}

\newpage

\section{Experimental Details}

\subsection{Architecture Details} \label{sec:ap_architecture_details}
In this section, we describe the neural network used to parameterize $\epsilon_\theta$. We adapted Graph Transformer (GT) from \url{https://github.com/lucidrains/graph-transformer-pytorch}. We modified the architecture to satisfy the symmetry constraints from Section 3.3. We start by naming a function that calls the Graph Transformer from the mentioned link as
\begin{equation*}
\mathrm{nodes}_\text{out} =  \mathrm{GT}[\mathrm{nodes}_\text{in}, \mathrm{edges}_\text{in}],    
\end{equation*}
where $\mathrm{nodes}_\text{in} \in \mathbb{R}^{n \times d_n}$, $\mathrm{edges}_\text{in} \in \mathbb{R}^{n \times n \times d_e}$. $n$ being the number of nodes, $d_n$ the number of dimensions per node and $d_e$ the number of dimensions per edge. $\mathrm{GT}$ will receive as input the node embeddings $\rmh \in \mathbb{R}^{n \times (\cdot)}$ and the noise level index $i$ as node features. Pairs of vector differences $\cg_j - \cg_k$ as edge features such that
\begin{align*}
\mathrm{nodes}_\text{in}[j,:] &= \mathrm{concat}\left[\rmh[j,:], i]\right]\\
\mathrm{edges}_\text{in}[j,k,:] &= \cg[j,:] - \cg[k,:]
\end{align*}

Now we define a network $\mathrm{nn}_{\theta'}: [\rmh, \cg, i] \rightarrow \mathbb{R}^1$ that takes the Graph Transformer and outputs a scalar value:

$$\mathrm{nn}_{\theta'}: \{\rmh, \cg, i\} \rightarrow \{\mathrm{nodes}_\text{in}, \mathrm{edges}_\text{in}\} \rightarrow \mathrm{GT} \rightarrow \{\mathrm{nodes}_\text{out}\} \rightarrow \mathrm{nn.Linear}(d_n, 1) \rightarrow \mathrm{sum}(\cdot) \rightarrow \{
\text{output}\} $$

Finally, to define $\epsilon_\theta$, we incorporate $\rmh$ as its learnable parameters $\theta = \{\theta', \rmh\}$ and compute the gradient of $\mathrm{nn}_{\theta'}$ w.r.t. $\cg$
$$
\epsilon_\theta(\cg, i) =\nabla_\cg \mathrm{nn}_{\theta'}(\rmh, \cg, i)
$$

\subsection{Coarse-graining Operator}
\begin{figure}[ht]
    \centering
    \includegraphics[width=1.0\textwidth]{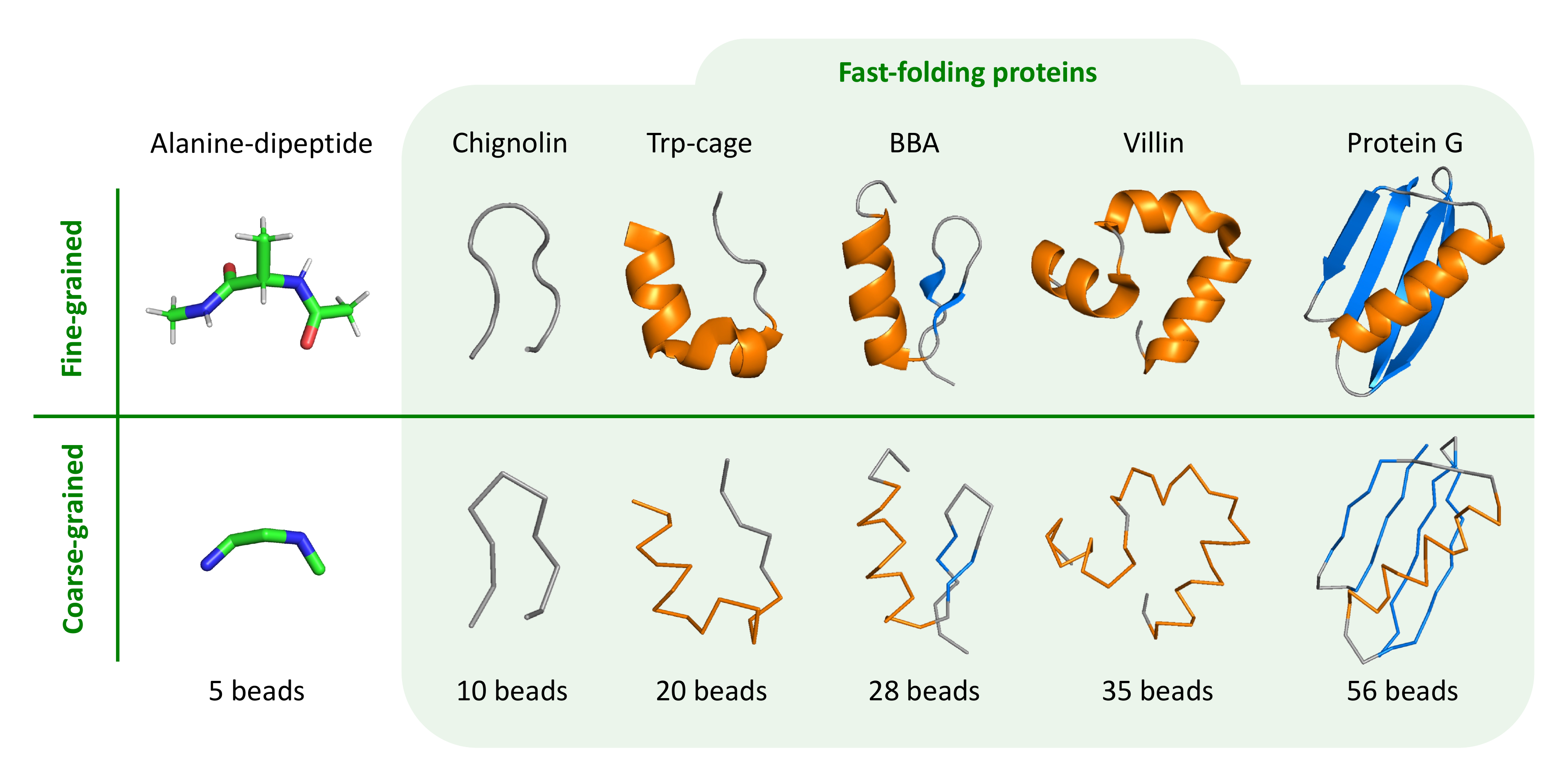}
    \caption{Overview of all systems used in our experiments, consisting of alanine dipeptide and five different fast-folding proteins. Top: fine-grained representation. Bottom: coarse-grained representation, i.e. backbone for alanine-dipeptide and $\rm{C_\alpha}$ atoms for the fast-folding proteins. 
    $\rm{\alpha}$-helices are shown in orange, $\rm{\beta}$-sheets are shown in blue.}
    \label{fig:data_figure}
\end{figure}

In \cref{fig:data_figure}, we give an overview of all systems we use in our experiments along with their CG representations. Coarse-graining is done by slicing out the relevant atoms, \ie the backbone atoms for alanine-dipeptide, and $\rm{C_{\alpha}}$ atoms for all fast-folding proteins.

\subsection{The optimization objective}
We defined the optimization objective of a diffusion model at a given noise level $i$ in Section 3 as $\E_{q(\cg_0)}\E_{\mathcal{N(\epsilon;\mathbf 0 , \mathbf I)}} \left[ \norm{\epsilon \!- \!\epsilon_\theta(\sqrt{\bar{\alpha}_i}\cg_0 + \sqrt{1  -  \bar{\alpha}_i}\epsilon, i)}^2 \right]$. In \citet{ho2020denoising}, the noise level $i$ is sampled from a discrete uniform distribution $i \sim \mathcal{U}\{1, \dots, L\}$ . In this work, in order to encourage the training at lower noise levels, we sample $i$ from $\mathcal{U}\{1, \dots, L/10\}$ and $\mathcal{U}\{L/10 + 1, \dots, L\}$ with $50 \%$ probability each. This results in more training samples at low noise levels which led to a better performance in the extracted force field.

\subsection{Alanine Dipeptide}
\label{sec:alanine_dipeptide}
The alanine dipeptide dataset simulated by \citet{kohler2022force} consists of four different simulations of length 500ns with 250K samples each. We partition the data using three simulations for train-validation (750K samples) and one for testing (250K samples). The three training-validation simulations are shuffled and 250K are used for validation and 500K for training.
\subsubsection{Implementation details} \label{ap:alanine_implementation_details}

To evaluate the influence of the dataset size on model performance, we train the model on the following amounts of training data: 10K, 20K, 50K, 100K, 200K and 500K.

We use a neural network with two graph Transformer layers and 96 hidden features per layer. The network is optimized using Adam with a learning rate of $3\cdot 10 ^{-4}$ and cosine learning rate decay dropping to $1\cdot 10 ^{-5}$. All experiments are run with batch size $1024$ and for $1$M iterations, with early stopping on the training set sizes $10$K and $20$K.

For the Langevin dynamics simulation, we use the same simulation settings as in \citet{kohler2022force}, except that we do not use parallel tempering. We simulate for $1$M iterations, with a time step resolution of $2$ femtoseconds, samples are saved every 250 steps resulting in 4K samples per simulation. 100 simulations are run in parallel resulting in a total of 400K samples. The mass of each CG node is set to $12.8$ g/mol, which is the weighted average of the mass of carbon and oxygen atoms. The temperature and the friction coefficient are the same as in the reference data (300 Kelvin and $1\text{ps}^{-1}$).

The noise level $i$ of the denoising diffusion process is obtained by cross-validation. As mentioned in Section 3.1, ideally, with infinite data and network capacity, we would choose $i$ to be the smallest value ($i=1$). But in practice we cross-validate it, and we find that the smaller the training dataset size, the larger the value of $i$. The $i$ values obtained for each amount of training  samples are the following (indexing from 0 as in the code instead of 1 as in the code):


\begin{table}[ht]
\setlength{\tabcolsep}{4pt}
\centering
\caption{Cross-validated noise levels $i$ across different training set sizes.}
\label{tab:noiselevels}
\vspace{4pt}
\begin{tabular}{@{}lcccccc@{}}
\toprule
Training set size & 10K & 20K & 50K & 100K & 200K & 500K \\ \midrule
Noise level $i$ & 26 & 25 & 20 & 19 & 17 & 8 \\ \bottomrule
\end{tabular}%
\end{table}

\subsection{Fast-folding Proteins} \label{sec:fastfolders}

\subsubsection{Dataset details} \label{ap:fastfolders_dataset_details}
All fast-folding protein data was obtained from \citet{lindorff2011fast}. The trajectories were randomly split 70-10-20\% into a training, validation and test set, using the same seed as \citet{kohler2022force}. All trajectories we used contain $\rm{C_{\alpha}}$ atoms only, and the interval between frames is 200 picoseconds. Further details can be found in the table below.


\begin{table}[h]
\setlength{\tabcolsep}{4pt}
\centering
\caption{Additional dataset details for fast-folding proteins.}
\label{tab:fast_folder_data_info}
\vspace{4pt}
\begin{tabular}{@{}lrrrrr@{}}
\toprule
 & \multicolumn{1}{c}{Chignolin} & \multicolumn{1}{c}{Trp-cage} & \multicolumn{1}{c}{Bba} & \multicolumn{1}{c}{Villin} & \multicolumn{1}{c}{Protein G} \\ \midrule
ID & \multicolumn{1}{c}{CLN025} & \multicolumn{1}{c}{2JOF} & \multicolumn{1}{c}{1FME} & \multicolumn{1}{c}{2F4K} & \multicolumn{1}{c}{NuG2/1MIO} \\
Temperature ($K$) & 340 & 290 & 325 & 360 & 350 \\
Amino acids & 10 & 20 & 28 & 35 & 56 \\
Simulation length ($\mu s$) & 106 & 208 & 223 & 125 & 369 \\
Data points & $534\,743$ & $1\,044\,000$ & $1\,114\,545$ & $629\,907$ & $1\,849\,251$ \\ \bottomrule
\end{tabular}%
\end{table}

\subsubsection{Implementation details} 
\label{ap:fastfolders_implementation_details}
In diffusion-denoising process, we used the cosine scheduler with 1000 different noise levels $i$ which is the standard setting in \citet{ho2020denoising,hoogeboom2022equivariant}. We used the Adam optimizer. Remaining training hyperparameters are reported in Table \ref{tab:training_parameters_fastfolders}. Given the training settings from \cref{tab:training_parameters_fastfolders}, the training iterations per second were 17 it/s for chignolin on two V100 GPUs and 13 it/s, 9.8 it/s, 6.5 it/s, 5.9 it/s, for trp-cage, bba, villin and protein G respectively on four V100 GPUs each.


\begin{table}[h]
\centering
\setlength{\tabcolsep}{4pt}
\caption{Hyperparameters used in different experiments.}
\label{tab:training_parameters_fastfolders}
\vspace{4pt}
\begin{tabular}{@{}lccccc@{}}
\toprule
 & Chignolin & Trp-cage & Bba & Villin & Protein G \\ \midrule
Batch size & 512 & 512 & 512 & 512 & 256 \\
Learning rate & $4\cdot10^{-4}$ & $4\cdot10^{-4}$ & $4\cdot10^{-4}$ & $4\cdot10^{-4}$ & $4\cdot10^{-4}$ \\
Training iterations & 1M & 1M & 2M & 2M & 3M \\
Number of layers & 3 & 3 & 3 & 3 & 3 \\
Number of features & 64 & 128 & 96 & 128 & 128 \\
Exponential moving average & .995 & .995 & .995 & .995 & .995 \\ \bottomrule
\end{tabular}%
\end{table}

For the Langevin dynamics, we ran 6 million steps simulations and saved samples every 500 steps, running 100 simulations in parallel resulted in 1.2M samples. The mass of the particles is set to 12 g/mol which is the mass of each slices Carbon atom, the simulated temperature is the same as in the ground truth data simulation which is reported in \cref{tab:fast_folder_data_info}). The noise levels $i$ used for each fast-folder protein indexing from $0$ are \{20, 15, 5, 5, 5\} for Chignolin, Trp-cage, Bba, Villin and Protein G respectively.

\subsubsection{Extended results}
\label{ap:fastfolders_extra}

\begin{figure*}[ht]
    \centering
    \includegraphics[width=1.0\textwidth]{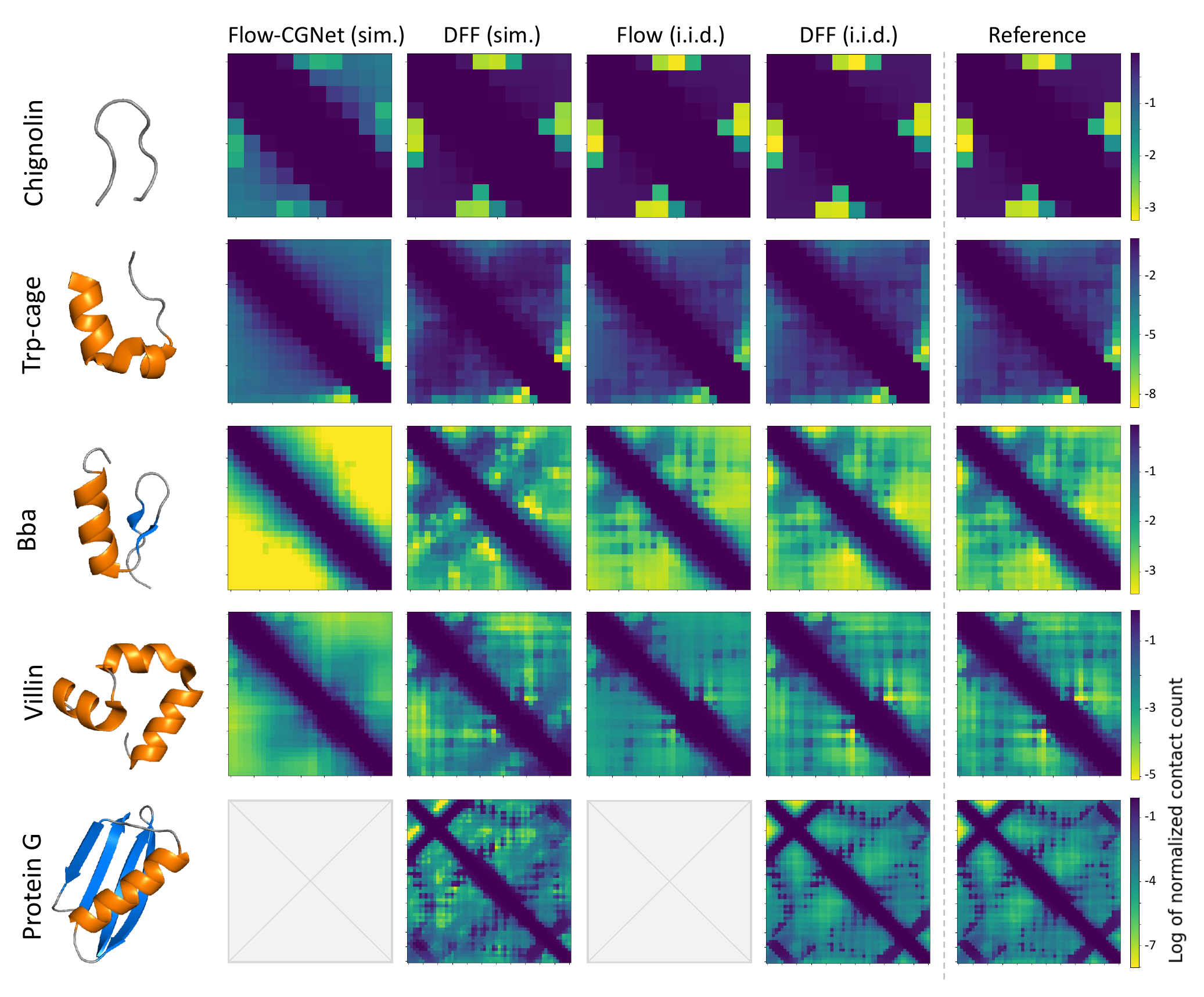}
    \caption{Contact analysis at equilibrium distribution for fast-folding proteins. Each row of results corresponds to one protein. Left: native structure visualization ($\rm{\alpha}$-helices in orange, $\rm{\beta}$-sheets in blue), made with PyMOL. Right: log of normalized contact counts over all samples. Red squares indicate region of interest that contains longer-range contacts, crucial toe global structure.}
    \label{fig:ap_contacts}
\end{figure*}

\paragraph{Contact maps} As presented in the main text, one way to analyze the global structure of a protein is by evaluating contact maps. Contacts are places within a protein where two atoms are close together, which will occur either when atoms are close in sequence space or when the protein's global fold demands the atoms to be within short distance range. These contacts can be shown as binarized pairwise distance matrices, where only those atoms pairs that are within a certain distance from one another are given value 1 while all other pairs are 0. Contact maps provide rich information about the global conformation of a protein, and they evolve over time during dynamics. Here, we choose to evaluate the distribution over contact maps at equilibrium. We get contact maps for all samples, where a pairwise distance between two atoms smaller than 10Å is considered a contact, and show a normalized count over these contact maps. Note that $\rm{C_{\alpha}}$ atoms are typically about 3.8Å apart due to local constraints, which means that the diagonal of the pairwise distance matrix as well as diagonals up until an offset between two and four are always contacts. From \cref{fig:ap_contacts}, it is clear that the DFF i.i.d. method is better at capturing the off-diagonal contacts, especially for larger structures. Even for our largest test case of Protein G, where we have no Flow-CGNet samples to compare, DFF i.i.d. extremely close to the reference. DFF sim. captures the off-diagonal distributions less well compared to the i.i.d. generator, but still improves significantly upon Flow-CGNet.

\begin{figure*}[ht]
    \centering
    \includegraphics[width=0.95\textwidth]{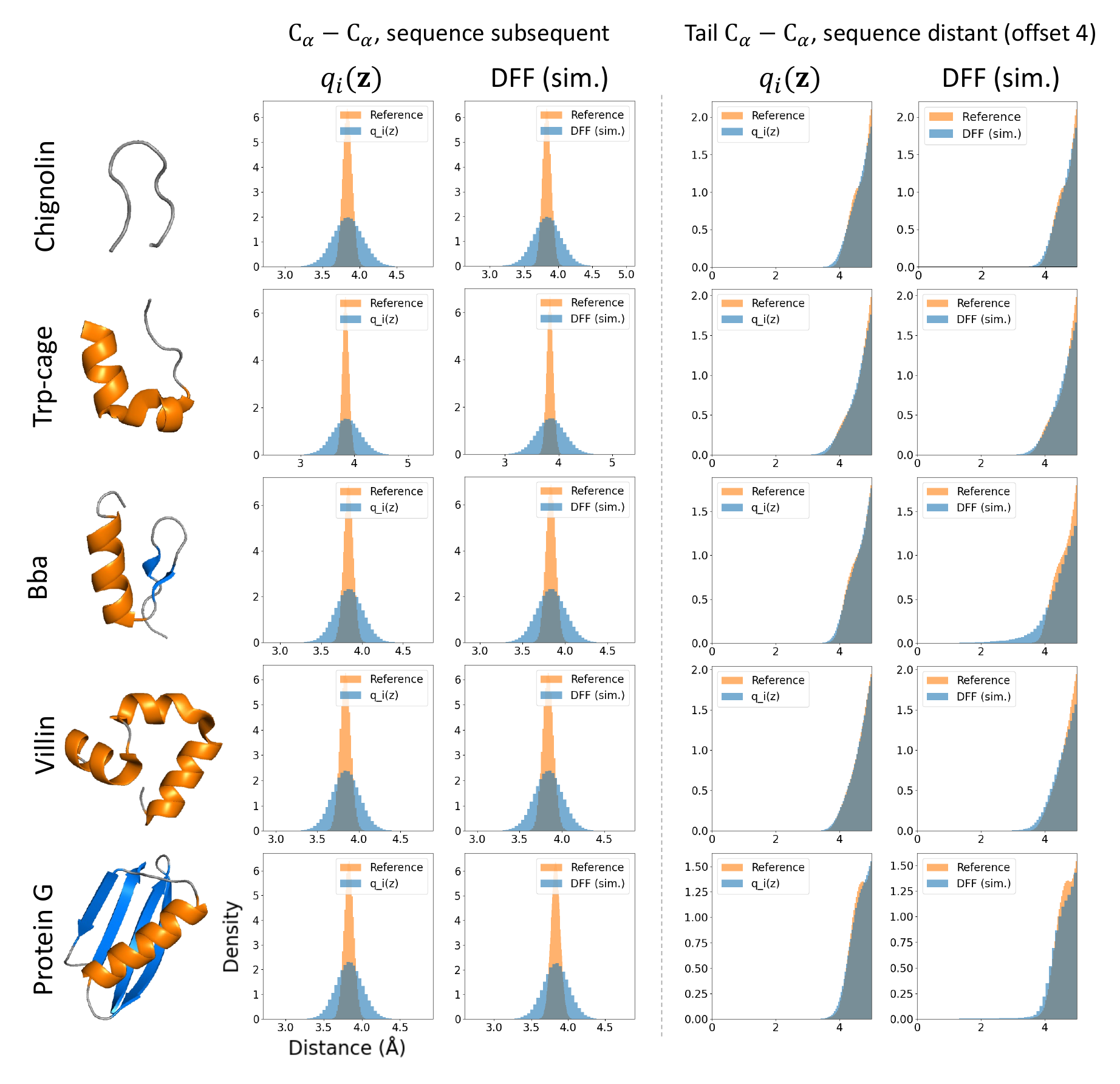}
    \caption{Chemical integrity analysis. Plots in the first two columns display the bond distribution of $q_i(\cg_i)$ and \textit{DFF (sim)} w.r.t. to \textit{Reference} Molecular Dynamics. This demonstrates that our model respects the bond distribution of $q_i(\cg)$ without the dynamics degenerating into unkown regions of space that would lead to bond dissociations. Plots in the right two columns display the small-distance tail of the distribution for sequence-distant atoms, these plots show there are no crossings of the backbone with itself (the distribution doesn't get close to zero).}
    \label{fig:chemical_integrity}
\end{figure*}

\paragraph{Chemical integrity} In order to show the chemical integrity of our samples, we show statistics over sequence-subsequent and sequence-distant $\rm{C_{\alpha}}$ atoms. In the coarse-grained molecules, $\rm{C_{\alpha}}$ atoms form the backbone of the protein structure, where we expect the bond length distribution between subsequent atoms to be consistent with the \textit{Reference} data through the simulation. In \cref{fig:chemical_integrity}, we compare the ground truth MD distribution to the samples from our \textit{DFF sim.} model. Our model shows the same mean, but a broader variance distribution due to the intrinsic variance introduced by the diffusion noise in $q_i(\cg)$. This is because \textit{DFF sim.} is actually modelling $q_i(\cg)$ to approximate the Reference data. To demonstrate that the variance increase in the bonded nodes is caused by the variance in $q_i(\cg)$ and not by a bond dissociation in the dynamics, we also plot the distribution of the ground truth $q_i(\cg)$, which we can sample by diffusing the reference MD data $\cg \sim q(\cg_i \mid  \cg_0)q(\cg_0)$. In \cref{fig:chemical_integrity}, we see the bonds distribution approximated by \textit{DFF (sim.)} are the same as the ground truth diffused distribution $q_i(\cg)$, this means, the dynamics do not diverge into unknown regions that would lead to bond dissociation.

Secondly, in the same \cref{fig:chemical_integrity}, we evaluate the small-distance tail of sequence-distant atoms that are three ore more residues apart, corresponding to off-diagonal contact points (see previous paragraph: "Contact maps"). When these small-distance tails contain values that approach zero, it means van der Waals forces are violated and in the worst case there can be crossings of the backbone with itself. \cref{fig:chemical_integrity} shows that our samples don't violate chemical integrity. 

\begin{figure*}[ht]
    \centering
    \includegraphics[width=1.0\textwidth]{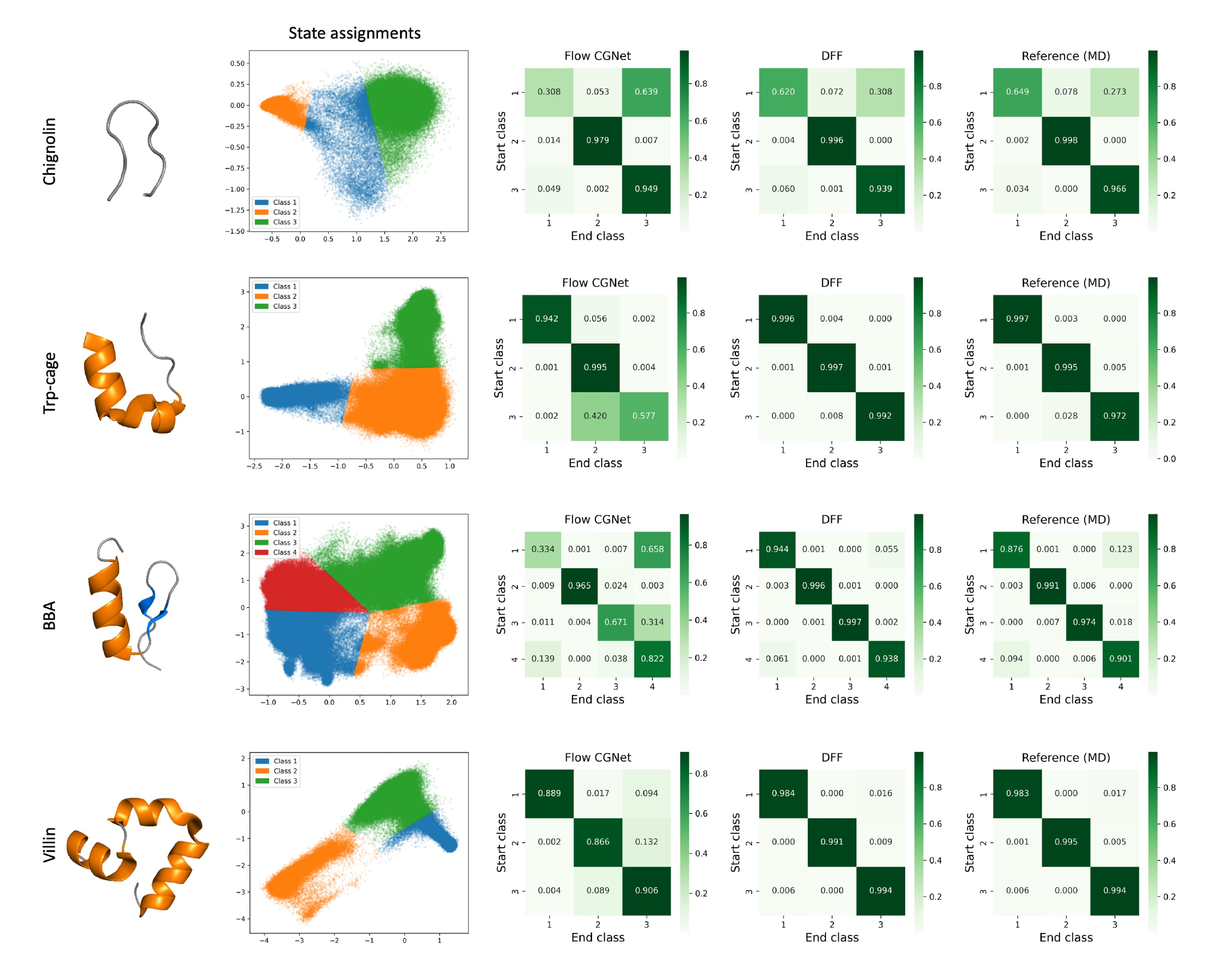}
    \caption{K-means clustering in 2D TIC space for fast-folders ($K$ determined by elbow method) and resulting transition probability matrices for Flow-CGNet, DFF and the reference distribution. Color intensity indicates probability.}
    \label{fig:ap_kmeans}
\end{figure*}

\paragraph{Transition probabilities using K-means clustering} In this paragraph we present more detailed results of transition probability analysis using K-means clustering. \cref{fig:ap_kmeans} depicts the results of k-means clustering in 2D TIC space as well as transition pribability matrices for the reference distribution, the Flow-CGNet model and our own DFF model. 

\end{document}